\documentclass[authoryear, nopreprintline, 12pt, 3p]{elsarticle}

\date{}
\usepackage{amssymb}
\usepackage{amsthm}

\usepackage{listings}
\usepackage{siunitx}

\usepackage[noend]{algpseudocode}
\usepackage{algorithm}

\usepackage{tabularx}
\usepackage{bbm}
\usepackage{bm}
\usepackage{multirow}
\usepackage{boldline}

\usepackage{subfig}

\usepackage{xcolor}

\usepackage[hidelinks]{hyperref}

\usepackage{amsmath}
\DeclareMathOperator{\argmin}{arg\,min} 

\algdef{SE}[SUBALG]{Indent}{EndIndent}{}{\algorithmicend\ }%
\algtext*{Indent}
\algtext*{EndIndent}

\journal{Neural Networks}

\begin{document}

\begin{frontmatter}



\title{Contrastive encoder pre-training--based \\ clustered federated learning for heterogeneous data}



\author[label1]{Ye Lin Tun}
\ead{yelintun@khu.ac.kr}
\author[label2]{Minh N.H. Nguyen}
\ead{nhnminh@vku.udn.vn}
\author[label1]{Chu Myaet Thwal}
\ead{chumyaet@khu.ac.kr}
\author[label1]{Jinwoo Choi}
\ead{jinwoochoi@khu.ac.kr}
\author[label1]{\\Choong Seon Hong \corref{cor1}}
\ead{cshong@khu.ac.kr}

\affiliation[label1]{organization={Department of Computer Science and Engineering, Kyung Hee University},
            city={Yongin-si},
            state={Gyeonggi-do 17104},
            country={South Korea}}
\affiliation[label2]{organization={Vietnam - Korea University of Information and Communication Technology},
            city={Danang},
            country={Vietnam}}
\cortext[cor1]{Corresponding author}


\begin{abstract}
\thispagestyle{empty}
{
Federated learning (FL) is a promising approach that enables distributed clients to collaboratively train a global model while preserving their data privacy.
However, FL often suffers from data heterogeneity problems, which can significantly affect its performance.
To address this, clustered federated learning (CFL) has been proposed to construct personalized models for different client clusters.
One effective client clustering strategy is to allow clients to choose their own local models from a model pool based on their performance.
However, without pre-trained model parameters, such a strategy is prone to clustering failure, in which all clients choose the same model.
Unfortunately, collecting a large amount of labeled data for pre-training can be costly and impractical in distributed environments.
To overcome this challenge, we leverage self-supervised contrastive learning to exploit unlabeled data for the pre-training of FL systems.
Together, self-supervised pre-training and client clustering can be crucial components for tackling the data heterogeneity issues of FL.
Leveraging these two crucial strategies, we propose contrastive pre-training--based clustered federated learning (CP-CFL) to improve the model convergence and overall performance of FL systems.
In this work, we demonstrate the effectiveness of CP-CFL through extensive experiments in heterogeneous FL settings, and present various interesting observations.
}

\end{abstract}


\end{frontmatter}


\section{Introduction}
\label{sec:introdution}

Training a neural network for intelligent applications demands a diverse and rich source of real-world data generated by end-user devices. These devices can range from personal smartphones to institutional data silos, where data privacy and security concerns are prominent \citep{10.5555/3152676}.
Conventionally, neural networks are trained on a central server with use of a collected dataset.
However, growing data privacy and security concerns may hinder data collection by service providers  \citep{10.5555/3152676}.
Moreover, transmitting large amounts of data from edge devices onto a central server can lead to high communication costs \citep{dinh2020federated, nguyen2020toward}.

To address these limitations, centralized learning systems are gradually transitioning towards federated systems, which offer better data privacy guarantees.
Edge devices in a federated learning (FL) system \citep{konevcny2016federated_2, konevcny2016federated, mcmahan2017communication} can collaboratively train a model without sharing sensitive data.
An FL system consists of a central server, coordinating the training, and a set of distributed client devices.
The training data reside on the client devices, and local models are directly trained by the clients.
The trained parameters are then transmitted to the server, while maintaining the confidentiality of each client's data. The server aggregates the received local parameters into a global model, which inherits the predictive capabilities of the local models.
As such, FL is an appealing option for services requiring training on privacy-sensitive data, such as medical data in the healthcare sector \citep{Kaissis2020SecurePA, 9373264, 9268161} or personal data on a smartphone \citep{hard2018federated, ramaswamy2019federated}.

Data heterogeneity poses a significant challenge in FL, as private data generated by each client device differ depending on the nature of the device \citep{Li2020FederatedLC, mcmahan2017communication, 8889996, yu2020heterogeneous, zhao2018federated}.
Non-independent and non-identically distributed (Non-IID) data generated by a large number of clients may cause the divergence between local parameter updates, thereby slowing the convergence speed and hurting the overall performance of the aggregated global model \citep{yu2020heterogeneous}
To address this challenge, clients can be partitioned into different clusters to train cluster-level models.
This kind of approach is often referred to as ``clustered federated learning'' (CFL) \citep{ghosh2020efficient, sattler2020byzantine, 9174890, 9174245}, where multiple cluster-level models are trained, as opposed to a single global model in conventional FL.
Depending on the clustering criteria, clients within the same cluster may share similar characteristics, such as data distributions or the availability of training resources.
Cluster models are trained by aggregating the local parameter updates generated by clients in the same cluster.
By taking advantage of the similar learning characteristics shared by cluster members, cluster models can deliver better personalized performance, such as higher accuracy for the local classification task.

Clustering in an FL environment poses complex challenges as client information cannot be directly accessed because of privacy constraints.
The issue of client clustering while adhering to FL constraints has attracted considerable attention recently \citep{briggs2020federated, duan2021fedgroup, ghosh2019robust, ghosh2020efficient, long2023multi, mansour2020three,  9174890}.
The iterative federated clustering algorithm (IFCA)~\citep{ghosh2020efficient} and HypCluster~\citep{mansour2020three} adopt a performance-based model selection approach where clients determine the cluster identities themselves.
Such an approach maintains multiple cluster models, and each client selects the model that performs best on its local data.
For instance, clients can measure the performance based on local learning loss or accuracy in the classification task.
Clients that choose the same model are considered members of the same cluster.
Allowing clients to choose their own model is a direct and intuitive way to improve the personalized performance, which is the main goal of CFL.

However, one problem with such an approach is that if cluster models are randomly initialized at the start, a particular random model may outperform all others. This may lead all clients to choose the same model, and therefore the clustering process may fail.
IFCA~\citep{ghosh2020efficient} assumes a good initialization where an initial cluster model $\theta^n$ is close to $\theta^{n*}$ to prove its convergence without clustering failure. IFCA also shows that it can succeed if the initialization requirements are relaxed with random initialization and multiple restarts.
Another potential approach is to reduce the randomness of initialized parameters by pre-training cluster models before the CFL process.
It is more practical compared with finding
a good initialization, or restarting the training with different sets of random parameters in the case of clustering failure.
However, obtaining labeled data is the most challenging aspect of pre-training, as it involves significant expenditure to gather and annotate the data for target tasks.
In some distributed environments, such data may not even be collectible because of privacy concerns.

On the other hand, unlabeled data with relevant characteristics and features as the target tasks may be widely available from public sources and the Internet.
Such unlabeled data may be subject to fewer privacy regulations and can be collected at a lower cost for pre-training.
Recently, self-supervised contrastive learning algorithms~\citep{bachman2019learning, caron2020unsupervised, chen2020simple, chen2021exploring, grill2020bootstrap, he2020momentum, misra2020self, zbontar2021barlow} have attracted major attention for leveraging unlabeled data to aid model training.
These algorithms pre-train the encoder part of the model to extract meaningful representations from raw data.
The pre-trained encoder increases the efficiency of labeled data when the model is fine-tuned for actual downstream tasks.
Contrastive learning assumes that two augmented instances originating from the same data sample should share similar information.
Accordingly, representations extracted from these two instances by the encoder should also be similar.
On the basis of this idea, the encoder is trained by minimizing the distance between two output representations in the embedding space \citep{bachman2019learning, chen2020simple, chen2021exploring, grill2020bootstrap, zbontar2021barlow}.

Since contrastive encoder pre-training and client clustering can be complementary solutions for providing an efficient design for practical FL systems, it is crucial that we explore how to effectively deploy these approaches together.
Therefore, in this study, we leverage contrastive learning to pre-train an encoder in a centralized setting with unlabeled data.
The pre-trained encoder is then deployed to enhance the CFL task, where we follow the same approach as in IFCA~\citep{ghosh2020efficient} for clustering the clients.
The goal is to investigate if and how these two prominent approaches can be effectively integrated to improve the model performance in a heterogeneous FL environment.
Our method is referred to as ``contrastive pre-training--based clustered federated learning'' (CP-CFL), and our contributions are summarized as follows:
\begin{itemize}
    \item We show that CP-CFL significantly improves the performance of CFL by leveraging contrastive encoder pre-training.

    \item We conduct extensive experiments using multiple pre-training datasets and downstream FL client datasets to evaluate the performance of CP-CFL. We explore efficient ways to deploy CP-CFL in the context of the pre-trained encoder and downstream classifier head. (More details are presented in Sections~\ref{sec:classifier_head} and \ref{sec:encoder_training}.) Furthermore, we provide various ablation studies to validate its effectiveness.

    \item Overall, our study contributes to a better understanding of how contrastive encoder pre-training and client clustering can jointly improve the performance of FL.

\end{itemize}

\section{Related work}
\label{sec:related_work}

In this section, we briefly discuss the FL process before covering various related studies on CFL and self-supervised contrastive learning.

\subsection{Federated learning and client clustering}
\label{sec:fl_and_client_clustering}

The core of FL is to allow distributed clients to collaboratively train a model while considering communication and privacy constraints \citep{Li2020FederatedLC, konevcny2016federated_2, konevcny2016federated, mcmahan2017communication}.
While conventional distributed training retains control over the training data, FL treats the data in each client device as private data. Such data cannot be shared with different parties such as other clients or the central server.
FL avoids the transmission of private data by directly training a model on each client while sharing only the trained model parameters with the server.
The server has no control over the clients, and a client may or may not participate in a training round and may even drop out of an ongoing one.
Usually, the client data in an FL environment are heterogeneous and follow different distributions dependent on the nature of the client \citep{hard2018federated, ramaswamy2019federated}.
However, vanilla FL is not specifically designed to handle the non-IID data of clients and trains a single global model for all clients, leading to unsatisfactory performance in a highly heterogeneous environment \citep{8889996, yu2020heterogeneous, zhao2018federated}.

One way to improve the performance in FL is to cluster the clients with similar learning characteristics and maintain separate models for each cluster.
Many approaches cluster the local parameters received at the server side based on different strategies, such as recursive bipartitioning \citep{duan2021fedgroup, 9174890}, task relatedness \citep{9739132}, cosine distance \citep{9774841}, stochastic expectation maximization \citep{long2023multi}, and agglomerative clustering \citep{briggs2020federated}.
\cite{li2021federated} performed soft clustering that allows clusters with overlapping clients, while \cite{dennis2021heterogeneity} proposed a one-shot scheme to perform clustering in a single FL round.
In addition, some studies have investigated the effectiveness of CFL in the presence of adversarial clients \citep{ghosh2019robust, sattler2020byzantine}. Other studies have used CFL for specific applications, such as handover prediction in wireless networks \citep{kim2021dynamic} and human activity recognition \citep{ouyang2022clusterfl}.
In contrast to the server-side client clustering strategy, IFCA~\citep{ghosh2020efficient} and HypCluster~\citep{mansour2020three} perform clustering on the client side by evaluating different models' performance on local data.

\subsection{Federated learning and contrastive learning}
\label{sec:fl_and_unsupervised_contrastive_learning}

Contrastive learning is a form of representation learning that exploits raw unlabeled data by training an encoder on the instance discrimination task.
This enables the encoder to learn generic representations that can be transferred to a wide variety of downstream tasks.
Unsupervised representation learning can be broadly categorized into two approaches: generative and discriminative.
Generative approaches typically learn representations by mapping the output pixels to the input pixels, such as autoencoders~\citep{kingma2013auto, vincent2008extracting}.
In contrast, discriminative approaches generate labels for a proxy task from the unlabeled data, and the encoder is trained on this proxy task \citep{gidaris2018unsupervised, noroozi2016unsupervised, pathak2016context}.
Contrastive learning~\citep{chen2020simple, chen2021exploring, grill2020bootstrap, he2020momentum} is one state-of-the-art discriminative approach that trains the encoder by minimizing the distance between representations of a positive pair (e.g., two different augmented views of the same image), while maximizing that of a negative pair (e.g., views from different images).
Besides the image domain, contrastive learning is also applied in video~\citep{dave2022tclr, pan2021videomoco, qian2021spatiotemporal}, audio~\citep{baevski2020wav2vec, manocha2021cdpam, saeed2021contrastive}, and natural language processing~\citep{gao2021simcse, giorgi2020declutr, wu2020clear}, making it suitable for FL environments that seek to provide a wide variety of services to clients.

Several studies have examined representation learning in the FL context, where contrastive learning is integrated into the client's local training step \citep{li2021model, van2020towards, wu2021federated, zhang2020federated, zhuang2021collaborative, zhuang2022divergence}.
FedU~\citep{zhuang2021collaborative}, FedCA~\citep{van2020towards}, and FedSSL~\citep{zhuang2022divergence} assume clients' private data are unlabeled and perform representation learning as a local training step.
RSCFed~\citep{liang2022rscfed} and FedCy~\citep{9950359} consider a semi-supervised setting where some clients contain labeled data and others contain unlabeled data.
Hetero-SSFL~\citep{makhija2022federated} allows clients to perform representation learning with different local model architectures.
FedCon~\citep{long2021fedcon} introduces a learning paradigm where the server contains labeled data and the clients contain unlabeled data.
MOON~\citep{li2021model}, FedCKA~\citep{son2022comparisons}, and FedIntR~\citep{tun2022federated} use contrastive learning to align the representations of global and local models.
In contrast, we use contrastive learning as a centralized pre-training step to construct a base encoder for CFL.
HotFed~\citep{9781151} and FedAUX~\citep{sattler2021fedaux} closely relate to our work since they pre-train the model using self-supervised contrastive learning; however, HotFed~\citep{9781151} transfers the pre-trained model to a vanilla FL setup, while FedAUX~\citep{sattler2021fedaux} transfers it to a federated distillation setting.
The aforementioned studies shed light on the possibility of incorporating contrastive pre-training into the CFL process, with the potential to increase the convergence rate and improve the performance of cluster models.

\begin{figure}[htb]
    \centering
    \includegraphics[width=\textwidth]{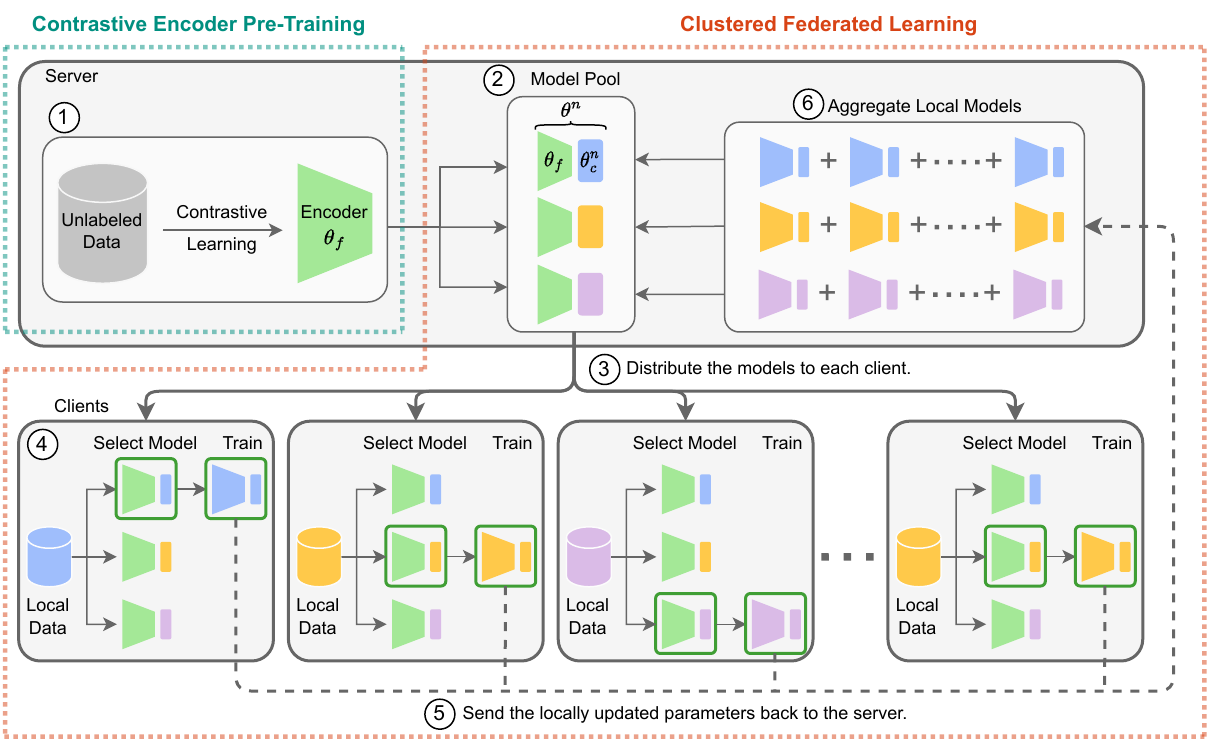}
    \caption{Overview of CP-CFL, which uses contrastive learning to pre-train the encoder $\theta_f$ on the unlabeled data. The pre-trained encoder $\theta_f$ is then deployed in the CFL task, which aims to train cluster-level models for different client clusters.}
    \label{fig:system_model}
\end{figure}

\section{Contrastive pre-training--based CFL}
\label{sec:contrastive_pretraining_based_cfl}

We present an overview of our CP-CFL framework before delving into details in later sections.

\subsection{Overview}

As illustrated in Figure~\ref{fig:system_model}, CP-CFL consists of two stages: the pre-training stage and the CFL stage.
In the pre-training stage, the central server uses contrastive learning to pre-train the encoder $\theta_f$ on unlabeled data.
To apply the pre-trained encoder $\theta_f$ in a useful downstream task such as classification, we can attach a classifier head $\theta_c$ to obtain a complete classification model $\theta = (\theta_f, \theta_c)$.
For a CFL task with $N$ clusters, the server generates a pool of $N$ classification models $\{\theta^n\}_{n=1}^N = \{(\theta_f, \theta_c^n)\}_{n=1}^N$  by joining the pre-trained encoder $\theta_f$ with $N$ randomly initialized classifier heads $\{\theta^n_c\}_{n=1}^N$.
In each global round, the model pool is distributed to the clients for clustering and local training.
Each client evaluates the models on its local dataset and selects the best-performing one with the lowest error.
The selected model is then trained, and updated parameters are sent back to the server.
Clients that choose the same model are identified as members of the same cluster.
Local parameter updates from the cluster members are aggregated to update the respective cluster model (i.e., the model mutually selected by the cluster members).
Each cluster model is tailored to the unique underlying data distribution of its respective cluster, providing better personalized performance to the clients.

\subsection{Problem formulation of CP-CFL}
\label{sec:problem_formulation_of_cpcfl}

\noindent
\textit{Contrastive pre-training}: Given an unlabeled dataset $\widetilde{\mathcal{D}}$, the encoder $\theta_f$ is trained to extract representations from each sample $\tilde{x} \in \widetilde{D}$.
These extracted representations $h=\theta_f(\tilde{x})$ should capture generic information transferable to a wide variety of downstream tasks.
To achieve this, we use a typical contrastive learning step for training. Firstly, the data sample $\tilde{x}$ is randomly augmented into two different instances, $x_i$ and $x_j$.
The encoder $\theta_f$ then extracts representations $h_i=\theta_f(x_i)$ and $h_j=\theta_f(x_j)$ from the augmented instances.
The goal is to train $\theta_f$ so that it generates similar $h_i$ and $h_j$.
Hence, a form of contrastive loss $\ell_{i,j}$ can be designed to minimize the distance between $h_i$ and $h_j$ in the embedding space.
(Section \ref{sec:encoder_pretraining_explain} provides further details on how $\ell_{i,j}$ is computed for different contrastive learning techniques.)
Generally, the encoder $\theta_f$ is trained by solving

\begin{equation}
    \min_{\theta_f} \frac{1}{|\widetilde{\mathcal{D}}|} \sum_{\tilde{x} \in \widetilde{\mathcal{D}}} \ell_{i,j} \text{.}
\label{eqn:encoder_training}
\end{equation}

\vspace{0.6em}
\noindent
\textit{Clustered federated learning}: Each FL client $u \in \mathcal{U}$ stores its own local private dataset $\mathcal{D}_u$, which can be used for model training.
However, a conventional FL system trains a single global model for all the clients in the population $\mathcal{U}$ without considering the heterogeneous nature of the clients.
In contrast, CFL assumes that clients in the population $\mathcal{U}$ can be further clustered into $N$ different groups (i.e., $\mathcal{G}^1, \mathcal{G}^2, \dots, \mathcal{G}^N$) based on their learning characteristics.
CFL aims to provide better personalized performance to the clients by fitting a cluster-level model $\theta^n$ for each cluster $\mathcal{G}^n$.
To take advantage of contrastive pre-training, each cluster model $\theta^n$ is constructed by combining the pre-trained encoder $\theta_f$ with a randomly initialized classifier head $\theta_c^n$.
We can train the cluster model $\theta^n$ for a cluster $\mathcal{G}^n$ by solving
\begin{equation}
    \min_{\theta^n} \frac{1}{|\mathcal{G}^n|} \sum\limits_{u \in \mathcal{G}^n} \mathcal{L}(\theta^n, \mathcal{D}_u),
\label{eqn:group_model}
\end{equation}
where $\mathcal{L}$ is the loss calculated for each client $u$ in the cluster $\mathcal{G}^n$. Specifically, we calculate  $\mathcal{L}$ as
\begin{equation}
    \mathcal{L}(\theta, \mathcal{D}) = \frac{1}{|\mathcal{D}|} \sum\limits_{(x,y)\in \mathcal{D}} \ell_{\text{CE}}(\theta, x, y),
\label{eqn:client_loss}
\end{equation}
where $(x, y)$ is the labeled data pair and $\ell_{\text{CE}}$ is the cross-entropy loss defined by
\begin{equation}
    \ell_{\text{CE}}(\theta, x, y) = - \sum_{m=1}^{|y|} y_m \log \theta(x)_m \text{,}
\label{eqn:ce_loss}
\end{equation}
where $y$ is assumed to be a one-hot encoded label vector, $|y|$ is the number of classes, and $y_m$ and $\theta(x)_m$ are the ground truth label and softmax probability prediction for the $m$th class, respectively.

Despite the advantages of CFL, determining the cluster structure $\{\mathcal{G}^n\}_{n=1}^N$ can be challenging, since we cannot access the clients' private information; therefore, we use the model selection strategy to determine the cluster identity $n_u \in \{1,\dots N\}$ of client $u$ without violating the FL constraints.
In this strategy, the server maintains a pool of cluster models $\{\theta^n\}_{n=1}^N$ that share the same architecture but different sets of parameters.
At each global round $t$, the server distributes all cluster models $\{\theta^n\}_{n=1}^N$ to the participating clients.
Each client $u$ evaluates $\{\theta^n\}_{n=1}^N$ on its local dataset $\mathcal{D}_u$ and then selects the best-performing model $\theta^{n_u}$ for its data distribution.
Specifically, clustering is achieved by solving Eq.~\eqref{eqn:cluster_identity} for each client $u$:

\begin{equation}
    n_u = \argmin_{n} \mathcal{L}(\theta^n, \mathcal{D}_u).
\label{eqn:cluster_identity}
\end{equation}
In Eq.~\eqref{eqn:cluster_identity}, $n \in \{1,\dots N\}$ and $n_u$ is the index of the selected model as well as the cluster identity of client $u$.
Each client $u$ locally trains the selected model $\theta^{n_u}$ on dataset $\mathcal{D}_u$, and then sends the updated parameters $\phi_u$ and the cluster identity $n_u$ back to the server.
Local parameter updates from the clients of the same cluster are aggregated by weighted averaging \citep{mcmahan2017communication}, shown in Eq. \eqref{eqn:weighted_avg}, to update the respective cluster model:
\begin{equation}
    \theta^n = \sum\limits_{u \in \mathcal{G}^{n}} \frac{|D_u|}{\sum\limits_{u' \in \mathcal{G}^{n}}|D_{u'}|} \phi_u .
\label{eqn:weighted_avg}
\end{equation}
In Eq. \eqref{eqn:weighted_avg}, $|D_u|$ refers to the size of the local dataset of client $u$.

\subsection{Contrastive encoder pre-training techniques}
\label{sec:encoder_pretraining_explain}

In our work, we explore multiple contrastive learning algorithms---namely, SimCLR~\citep{chen2020simple}, BYOL~\citep{grill2020bootstrap}, and SimSiam~\citep{chen2021exploring}---for pre-training the encoder.
We describe the common key components of these contrastive learning algorithms as follows:
\begin{itemize}
    \item \textit{Random augmentation}: Augmentations are necessary to generate a pair of positive samples from an input sample.
    For each sample $\tilde{x} \in \widetilde{\mathcal{D}}$, random augmentations are applied to generate a positive sample pair: $x_i=\text{Augment}(\tilde{x})$ and $x_j=\text{Augment}(\tilde{x})$.

    \item \textit{Encoder} $\theta_f$: Contrastive learning pre-trains the encoder $\theta_f$ to extract generic representations $h$ given an input $\tilde{x}$.
    The pre-trained encoder $\theta_f$ can be paired with a task-specific head (in our case, we attach a classifier head $\theta_c$) and fine-tuned on a small labeled dataset to perform a useful downstream task.

    \item \textit{Distance metric}: Many contrastive learning approaches use cosine similarity to measure the distance between two representations \citep{chen2020simple, chen2021exploring, grill2020bootstrap}. The cosine similarity between two representation vectors $v_i$ and $v_j$ is calculated as
    \begin{equation}
    \text{Sim}(v_i, v_j)=\frac{v_i}{{\lVert v_i \rVert}_2} \cdot \frac{v_j}{{\lVert v_j \rVert}_2},
    \label{eqn:cosine_similarity}
    \end{equation}
    where ${\lVert . \rVert}_2$ is the $\ell_2$-norm.

\end{itemize}

\begin{figure}[htbp]
    \centering
    \subfloat[SimCLR]{\includegraphics[width=0.325\textwidth]{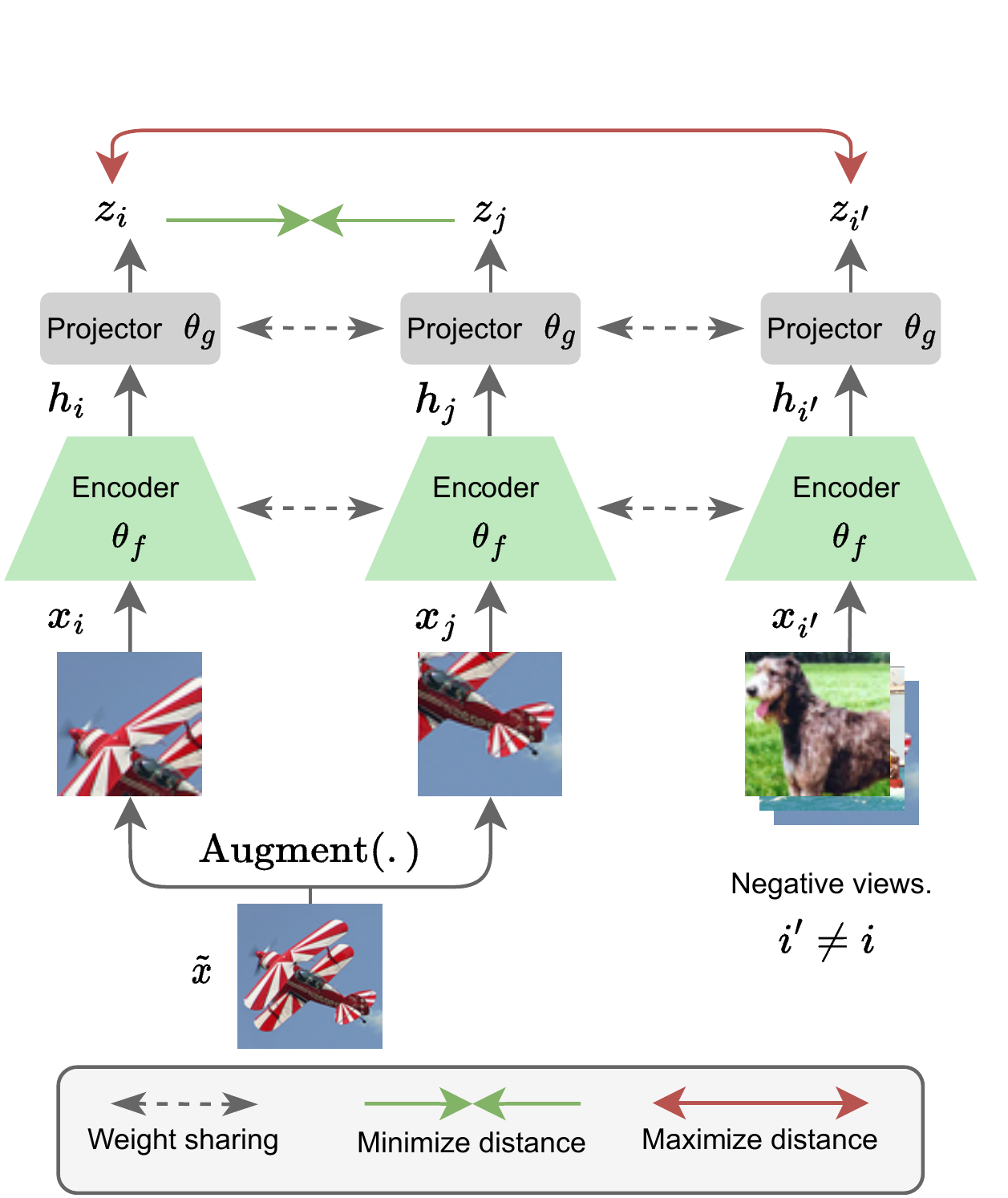}
     \label{subfig:url_simclr}}
	\subfloat[BYOL]{\includegraphics[width=0.325\textwidth]{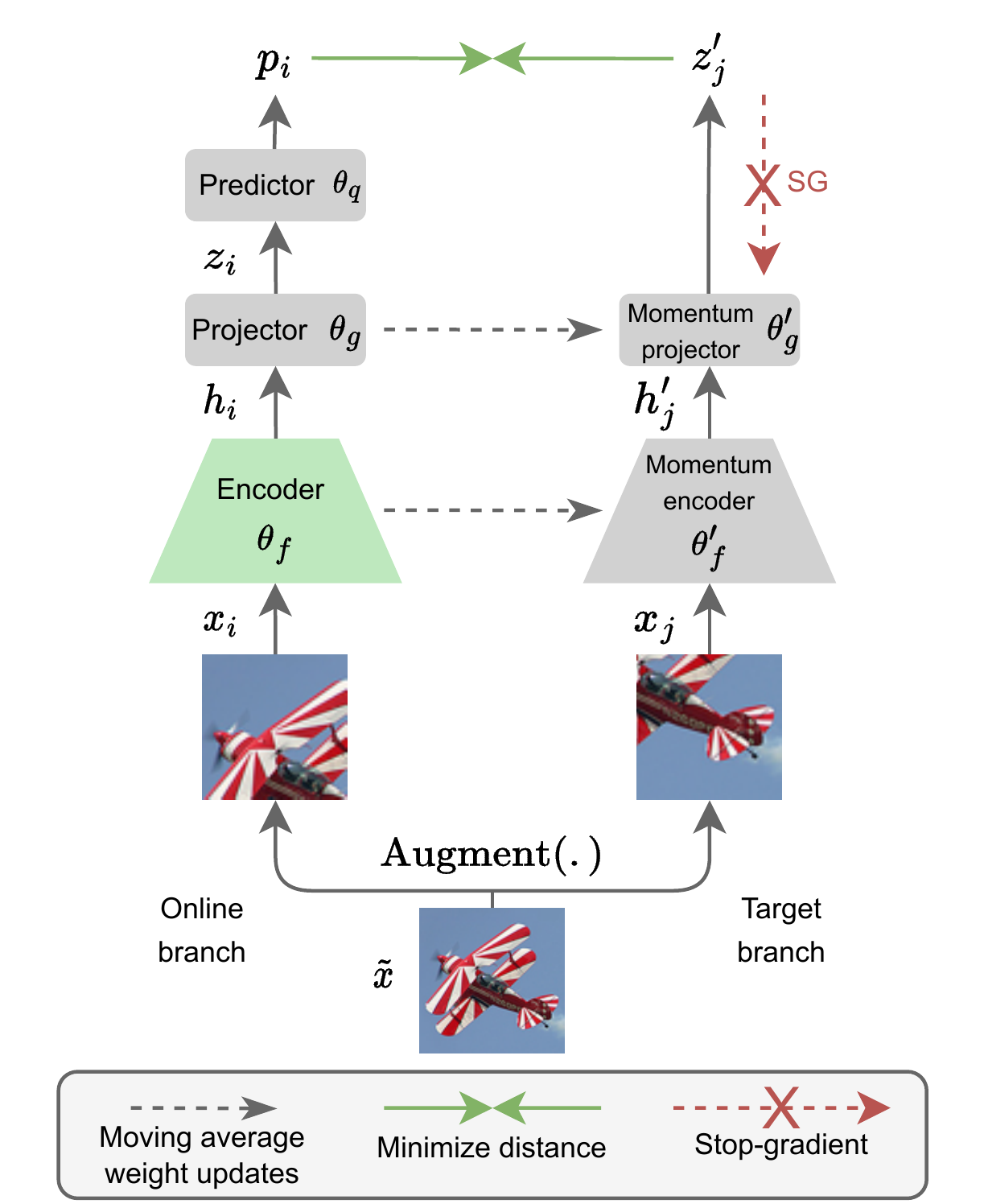}
 \label{subfig:url_byol}}
	\subfloat[SimSiam]{\includegraphics[width=0.325\textwidth]{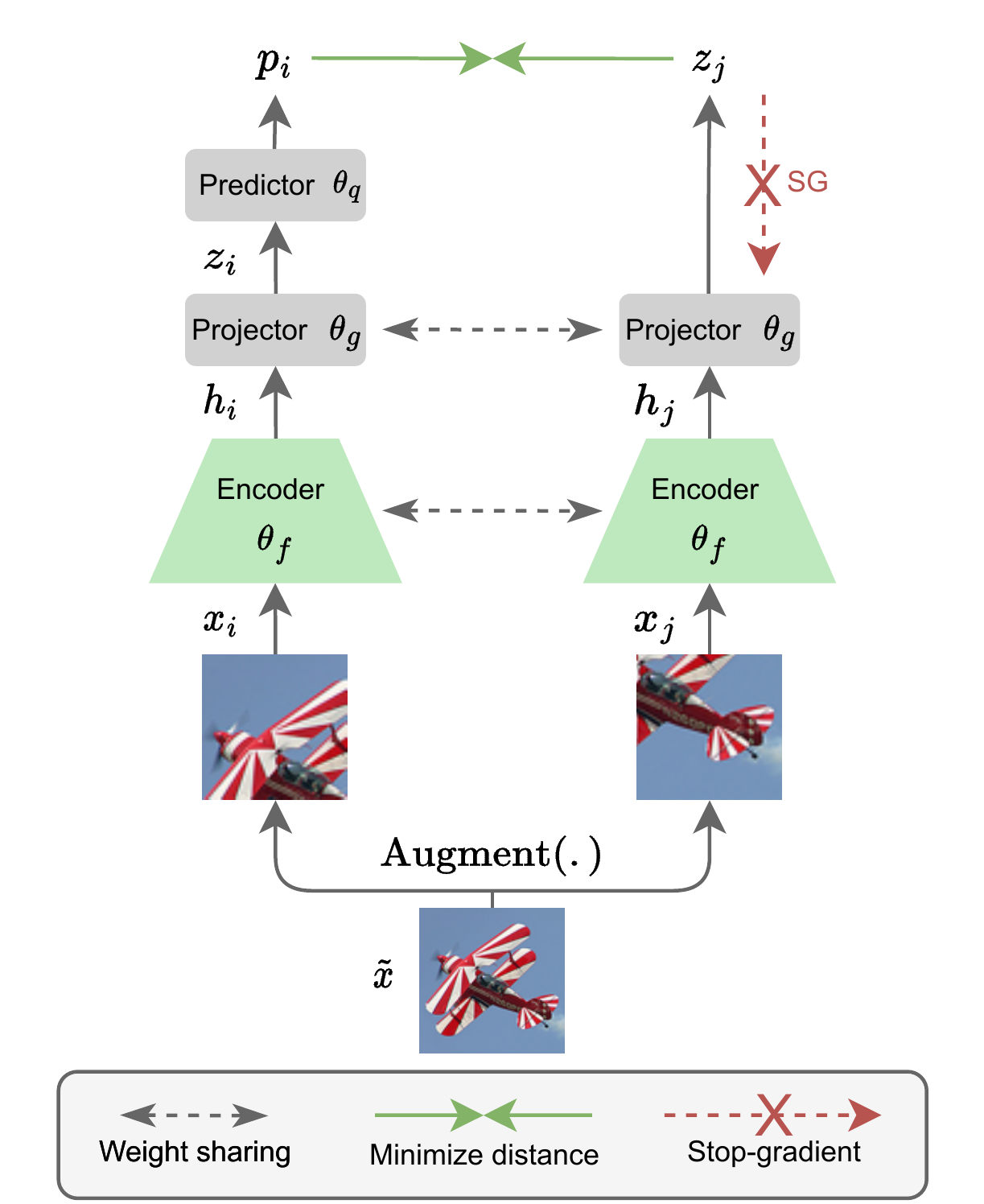}
 \label{subfig:url_simsiam}}
	\caption{Self-supervised contrastive learning techniques: (a) SimCLR \citep{chen2020simple}; (b) BYOL \citep{grill2020bootstrap}; (c) SimSiam \citep{chen2021exploring}.}
	\label{fig:url}
\end{figure}

In the following subsections, we provide a brief overview of each of the three contrastive learning approaches. Table~\ref{tab:different_url} and Figure~\ref{fig:url} compare the detailed characteristics of different contrastive learning approaches, while Algorithm~\ref{alg:url_pretraining} describes the general contrastive learning procedure.

\begin{table}[htbp]
\centering
\begin{tabular}{lV{3}cccccc}
\hlineB{3}
        & \begin{tabular}[c]{@{}c@{}}Positive\\ samples\end{tabular} & \begin{tabular}[c]{@{}c@{}}Negative\\ samples\end{tabular} & \begin{tabular}[c]{@{}c@{}}Weight\\ sharing\end{tabular} & Projector  & Predictor   & \begin{tabular}[c]{@{}c@{}}Stop\\ gradient\end{tabular}                  \\ \hlineB{3}
SimCLR  & \checkmark       & \checkmark       & \checkmark     & \checkmark    & \textendash   & \textendash   \\
BYOL    & \checkmark       & \textendash      & \textendash    & \checkmark    & \checkmark    & \checkmark    \\
SimSiam & \checkmark       & \textendash      & \checkmark     & \checkmark    & \checkmark    & \checkmark    \\ \hlineB{3}
\end{tabular}
\caption{Comparison between different self-supervised contrastive learning techniques.}
\label{tab:different_url}
\end{table}

\begin{algorithm}[tb]
\caption{Contrastive encoder pre-training}
\label{alg:url_pretraining}
\begin{algorithmic}[1]
    \State \textbf{Input:} encoder $\theta_f$, unlabeled dataset $\widetilde{\mathcal{D}}$, number of epochs $E$, learning rate $\eta$.
    \item[] 
    \State \textbf{Contrastive pre-training}($\theta_f$, $\widetilde{\mathcal{D}}$, $E$, $\eta$):
    \Indent
    \For{epoch $e = 0, 1, \dots, E-1 $}
        \For{ each batch $\mathcal{B} \in \widetilde{\mathcal{D}}$}
            \For{ each sample $\tilde{x} \in \mathcal{B}$}
                \State $x_i = \text{Augment}(\tilde{x})$
                \State $x_j = \text{Augment}(\tilde{x})$
                \State $h_i = \theta_f(x_i)$
                \State $h_j = \theta_f(x_j)$
                \State Calculate $\ell_{i,j}$ using $h_i$ and $h_j$ \Comment{Use Eq. \eqref{eqn:simclr}, \eqref{eqn:byol}, or \eqref{eqn:simsiam}.}
            \EndFor
            \State $\mathcal{L} = \frac{1}{|\mathcal{B}|} \sum\limits_{\tilde{x} \in \mathcal{B}} \ell_{i,j}$
            \State Update $\theta_f$ with $\eta\nabla\mathcal{L}$
        \EndFor
    \EndFor
    \EndIndent
    \item[] 
    \State \textbf{Output:} encoder $\theta_f$
\end{algorithmic}
\end{algorithm}

\subsubsection{SimCLR}
\label{sec:simclr}

For each image $\tilde{x}$ in an input batch $\mathcal{B}$, SimCLR \citep{chen2020simple} generates two views, $x_i$ and $x_j$, using a random augmentation function, $\text{Augment}(.)$. (This results in a total of $2|\mathcal{B}|$ augmented views for each input batch $\mathcal{B}$.) The two augmented views $x_i$ and $x_j$ are considered a positive pair since they originate from the same image $\tilde{x}$. The encoder $\theta_f$ extracts representations $h_i=\theta_f(x_i)$ and $h_j=\theta_f(x_j)$ from the augmented views, while the projection head $\theta_g$ transforms the extracted representations into projections $z_i=\theta_g(h_i)$ and $z_j=\theta_g(h_j)$. If we consider $x_i$ as the anchor sample, then $x_j$ acts as the corresponding positive counterpart, and vice versa. Using $x_i$ as the anchor, SimCLR calculates its loss based on the InfoNCE loss \citep{van2018representation} as
\begin{equation}
    \ell_{i,j} = - \log\frac{\exp(\text{Sim}(z_i, z_j)/\tau)}{\sum\limits_{i'=1}^{2|\mathcal{B}|} \mathbbm{1}_{[i' \neq i]} \exp(\text{Sim}(z_i, z_{i'})/\tau)}
    \label{eqn:simclr} \text{,}
\end{equation}
where $\tau$ is the temperature. Every other augmented view $x_{i'}$, where $i'=\{1, \dots, 2|\mathcal{B}|\}, i' \neq i$, is treated as a negative counterpart to $x_i$, with the distance between their representations maximized. These steps are repeated by treating each augmented view in a batch as an anchor. In SimCLR, negative samples help prevent the solution from collapsing into a trivial constant \citep{chen2021exploring}.

\subsubsection{BYOL}
\label{sec:byol}

BYOL~\citep{grill2020bootstrap} uses an online network $\theta$ branch  and a target network $\theta'$ branch for training.
During training, augmented views $x_i$ and $x_j$ are fed into the online and the target branches, respectively.
The online branch extracts $h_i=\theta_f(x_i)$ and $z_i=\theta_g(h_i)$, while the target branch uses the momentum encoder $\theta_f'$ and the momentum projector $\theta_g'$ to extract $h_j'=\theta_f'(x_j)$ and $z_j'=\theta_g'(h_j')$.
The momentum encoder $\theta_f'$ and the momentum projector $\theta_g'$ have the same structure as $\theta_f$ and $\theta_g$, but their parameters are moving-average versions of those of $\theta_f$ and $\theta_g$, respectively.
The online branch makes a prediction $p_i=\theta_q(z_i)$ using the predictor $\theta_q$ and the network is trained by minimizing the distance between $p_i$ and $z_j'$. In a sense, BYOL trains the encoder $\theta_f$ by using a view $x_i$ to predict the representation of the other view $x_j$. The loss between the anchor and its positive sample is computed as follows:
\begin{equation}
    \ell_{i,j} \triangleq {\lVert p_i - z_j'\rVert}_2^2 = 2 - 2 \cdot \text{Sim}(p_i, z_j') .
    \label{eqn:byol}
\end{equation}
BYOL also swaps the inputs $x_j$ and $x_i$ for each branch and then calculates the prediction $p_j$ and target branch output $z_i'$ to minimize the distance between them. Only the online branch is updated with gradient backpropagation, while the target branch parameters are updated as the moving-average versions of the online branch using Eqs. \eqref{eqn:byol_target_encoder} and \eqref{eqn:byol_target_projector}:
\begin{equation}
    \theta_f' = \beta \theta_f' + (1-\beta) \theta_f \text{,}
    \label{eqn:byol_target_encoder}
\end{equation}
\begin{equation}
    \theta_g' = \beta \theta_g' + (1-\beta) \theta_g \text{,}
    \label{eqn:byol_target_projector}
\end{equation}
where $\beta$ is the target branch update rate. BYOL uses the target branch and stops the gradient flow through it to prevent the solution from collapsing to a constant \citep{chen2021exploring}.

\subsubsection{SimSiam}
\label{sec:simsiam}

SimSiam~\citep{chen2021exploring} shares similar properties with both SimCLR and BYOL.
SimSiam uses the weight-sharing encoder $\theta_f$ and projector $\theta_g$ as in SimCLR to extract representations $h_i=\theta_f(x_i)$ and $h_j=\theta_f(x_j)$, as well as the projections $z_i=\theta_g(h_i)$ and $z_j=\theta_g(h_j)$. Similarly to BYOL, the predictor $\theta_q$ in one branch of SimSiam makes a prediction $p_i=\theta_q(z_i)$ to predict the output $z_j$ of the other branch. The contrastive loss is calculated as the negative cosine similarity between the prediction $p_i$ and the projection $z_j$:
\begin{equation}
    \ell_{i,j} = -\text{Sim}(p_i, z_j) \text{.}
    \label{eqn:simsiam}
\end{equation}
The inputs $x_j$ and $x_i$ can also be swapped for each branch, and the predictor $\theta_q$ can make the prediction $p_j$ to contrast with the output $z_i$ of the other branch. The encoder is trained by minimizing the distance between $p_i$ and $z_j$, as well as the distance between $p_j$ and $z_i$. SimSiam stops the gradient flow through the branch without the predictor to prevent the solution from collapsing to a constant.

\begin{algorithm}[htbp]
\caption{Contrastive pre-training--based clustered federated learning}
\label{alg:cfed}
\begin{algorithmic}[1]
    \State \textbf{Input:} encoder $\theta_f$, classifier head $\theta_c$, unlabeled dataset $\widetilde{\mathcal{D}}$, number of contrastive epochs $E_p$, contrastive learning rate $\eta_p$, number of clusters $N$, number of rounds $T$, number of classifier head training rounds~$T_c$, number of local epochs $E_\ell$, local learning rate $\eta_\ell$.
    \item[] 
    \State \textbf{Server executes:}
    \Indent
    \State $\theta_f =$ \textbf{Contrastive pre-training}($\theta_f$, $\widetilde{\mathcal{D}}$, $E_p$, $\eta_p$) using Algorithm~\ref{alg:url_pretraining}

    \For {$n=1, 2, \dots, N$}             \Comment{Generate a pool of $N$ cluster models.}
        \State $\theta_c^n =$ randomly initialize $\theta_c$
        \State Join $\theta_f$ and $\theta_c^n$ to obtain cluster model $\theta^n = (\theta_f, \theta_c^n)$
    \EndFor

    \For{round $t = 0, 1, \dots, T-1$}
        \State Initialize empty clusters $\mathcal{G}^n = \{\}$, where $n=\{1,\dots,N\}$
        \State Get participating client population $\mathcal{U}$
        \For{each client $u \in \mathcal{U}$ in parallel}
            \State $(\phi_u, n_u) =$ \textbf{Local update}($u$, $t$, $\{\theta^n\}_{n=1}^N$)
            \State Append $u$ to $\mathcal{G}^{n_u}$
        \EndFor
        \For {each cluster $\mathcal{G}^{n}$, where $n=1,2,\dots,N$},
            \State $\theta^n = \sum\limits_{u \in \mathcal{G}^{n}} \frac{|D_u|}{\sum\limits_{u' \in \mathcal{G}^{n}}|D_{u'}|} \phi_u$
        \EndFor
    \EndFor
    \EndIndent

    \item[] 
    \State \textbf{Clients execute:} \textbf{Local update}($u$, $t$, $\{\theta^n\}_{n=1}^N$):
    \Indent
    \If{$t<T_c$}                    \Comment{Explore with a random model.}
        \State $n_u = \text{select randomly from } \{1,2,\dots,N\}$
        \State Freeze the encoder part $\theta_f$ of $\theta^{n_u}$
    \Else                \Comment{Exploit the best model.}
        \State $n_u = \argmin_{n} \mathcal{L}(\theta^n, \mathcal{D}_u)$, where $n = 1, 2, \dots, N$
    \EndIf
    \State Synchronize local model $\phi_u = \theta^{n_u}$
    \For{epoch $e = 0, 1, \dots, E_\ell-1 $}
        \For{each batch $\mathcal{B} \in \mathcal{D}_u$}
            \State $\mathcal{L} = \frac{1}{|\mathcal{B}|} \sum\limits_{(x,y) \in \mathcal{B}} \ell_{\text{CE}}(\phi_u, x, y)$
            \State $\phi_u = \phi_u - \eta_\ell \nabla \mathcal{L}$
        \EndFor
    \EndFor
    \State return $(\phi_u, n_u)$ to server
    \EndIndent
    \item[] 
    \State \textbf{Output:} cluster models $\{\theta^n\}_{n=1}^N$, clusters $\{\mathcal{G}^n\}_{n=1}^N$
\end{algorithmic}
\end{algorithm}

\subsection{CP-CFL algorithm}
\label{sec:cpcfl_algorithm}

The detailed procedure of CP-CFL is presented in Algorithm~\ref{alg:cfed}.
Initially, the server pre-trains the encoder $\theta_f$ on an unlabeled dataset $\widetilde{\mathcal{D}}$ using contrastive learning.
The pre-trained encoder $\theta_f$ is then joined with a randomly initialized classifier head $\theta_c^n$ to generate a cluster model $\theta^n = (\theta_f, \theta_c^n)$, where $n=1,2,\dots,N$. Although the encoder $\theta_f$ is pre-trained, random parameters in the classifier head $\theta_c^n$ still have the potential to damage the clustering process. Therefore, for a few initial rounds (i.e., when $t<T_c$), the clients explore by selecting a model at random  rather than based on the performance. During this period, the encoder $\theta_f$ of each cluster model $\theta^n$ is frozen, with the local training updating only  the classifier head $\theta_c^n$.
Here, $T_c$ is significantly smaller than the total number of rounds $T$ (by default, we set $T_c=10$ and $T=100$). When $t \geq T_c$, each client $u$ determines its cluster identity $n_u$ by evaluating the cluster models  $\{\theta^n\}_{n=1}^N$ on the local dataset $\mathcal{D}_u$. Client $u$ then trains the entire model $\theta^{n_u}$ and then sends the locally updated parameters $\phi_u$ and its cluster identity $n_u$ back to the server. Local model updates from the clients of the same cluster $\mathcal{G}^n$ are aggregated to update the corresponding cluster model $\theta^n$.

\subsection{Characteristics of CP-CFL}
\label{sec:attributes_cpcfl}

To provide a clear insight into CP-CFL, we compare the characteristics of CP-CFL with other baseline approaches in Table~\ref{tab:cpcfl_attribute_comparison} .
In contrast to vanilla FedAvg~\citep{mcmahan2017communication}, which trains a single global model, IFCA~\citep{ghosh2020efficient} and CP-CFL train multiple cluster models.
Models in FedAvg and IFCA are randomly initialized, whereas CP-CFL pre-trains the encoder part with self-supervised contrastive learning.
IFCA relaxes the random initialization with multiple restarts to handle the clustering failure, while CP-CFL relies on the pre-trained encoder and $T_c$.
CP-CFL performs the pre-training on the server without extra computational burden on client devices. Moreover, $T_c$ in CP-CFL is usually small and can be integrated into the total number of rounds $T$, requiring no extra communication cost than IFCA.
Generally, the communication cost $C$ for an FL client can be calculated as follows \citep{qu2022rethinking}:
\begin{equation}
    C = T \times ( C_\text{sc} + C_\text{cs}),
    \label{eqn:comm_cost}
\end{equation}
where $T$ is the number of rounds, and $C_\text{sc}$ and $C_\text{cs}$ denote the server-to-client and client-to-server costs, where both can be derived as $ \text{number of transmitted models} \times \text{model size}$. We use the same model architecture in all approaches and denote the model size as $S$. Based on above formulation, we calculate the cost for FedAvg~\citep{mcmahan2017communication} as
\begin{equation}
    C_\text{FedAvg} = T \times (1 \times S + 1 \times S) = 2ST
    \label{eqn:comm_cost_fedavg}
\end{equation}
and the cost for CFL approaches as
\begin{equation}
    C_\text{CFL} = T \times (N \times S + 1 \times S) = (N+1)ST \text{.}
    \label{eqn:comm_cost_cfl}
\end{equation}
In general, CFL approaches are $\frac{N+1}{2}$ times more expensive than FedAvg in terms of communication cost. For instance, recalling that $N$ is the number of cluster models, we find that CFL approaches are twice as expensive than FedAvg when $N=3$.

\begin{table}[htbp]
\centering
\begin{tabular}{lV{3}c|c|c|c}
\hlineB{3}
       & \begin{tabular}[c]{@{}c@{}}Number of\\models\end{tabular} & \begin{tabular}[c]{@{}c@{}}Model\\initialization\end{tabular} & \begin{tabular}[c]{@{}c@{}}Handling\\clustering failure \end{tabular}     &  \begin{tabular}[c]{@{}c@{}}Communication\\cost\end{tabular} \\ \hlineB{3}
FedAvg & Single     & Random                                                               & -- & $2ST$ \\ \hline
IFCA   & Multiple   & Random                                                               & \begin{tabular}[c]{@{}c@{}} Multiple\\restarts\end{tabular}     & $(N+1)ST$ \\ \hline
CP-CFL & Multiple   & \begin{tabular}[c]{@{}c@{}}Contrastive\\ pre-training\end{tabular}   & \begin{tabular}{ccc}
\begin{tabular}[c]{@{}c@{}}Pre-trained\\ encoder\end{tabular} & + & $T_c$
\end{tabular}    & $(N+1)ST$ \\ \hlineB{3}
\end{tabular}
\caption{Characteristics of CP-CFL.}
\label{tab:cpcfl_attribute_comparison}
\end{table}

\section{Experiments}
\label{sec:experiment}

In this section, we present the evaluation results of CP-CFL on various experimental settings. We first describe the common experimental settings, before delving into the details of our studies.

\subsection{Experimental settings}
\label{sec:exp_settings}
\begin{figure}[htb]
    \centering
    \includegraphics[width=\textwidth]{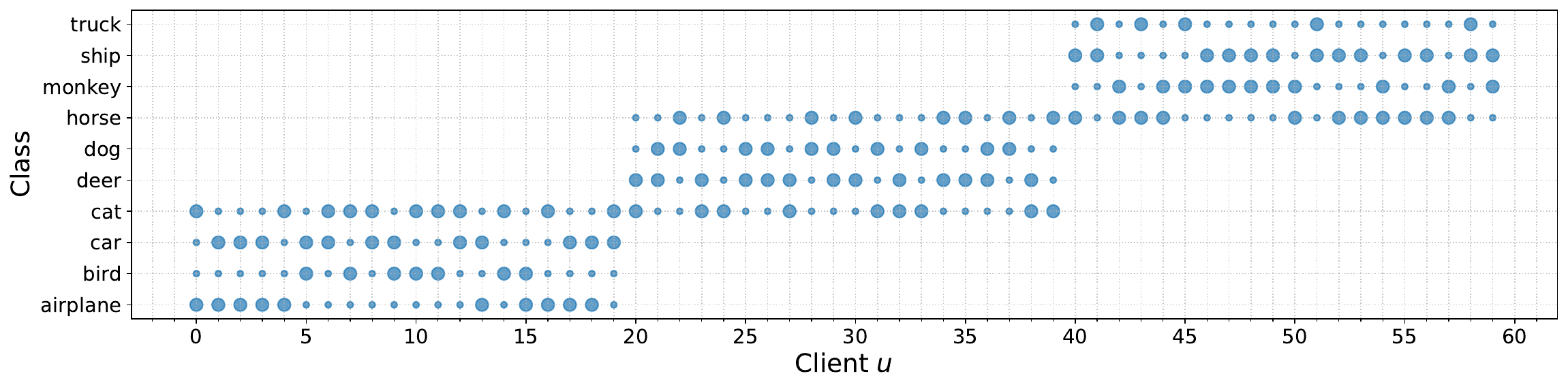}
    \caption{The amount of private data in clients for different classes, sampled from the training portion of the STL-10 dataset. Small dots and large dots represent $5$ and $20$ images, respectively.}
    \label{fig:client_data_amount}
\end{figure}

\noindent
Unless otherwise stated, we use the following settings in all of our experiments.
\vspace{0.6em}
\newline \textit{Dataset}: For our experiments, we mainly use the STL-10 dataset~\citep{coates2011analysis}, which contains both unlabeled and labeled images with a shape of $96 \times 96$. The unlabeled data portion contains $\num[group-separator={,}]{100000}$ images, whereas the labeled training and testing portions  contain $5000$ and $8000$ image-label pairs for $10$ different classes, respectively.
\vspace{0.6em}
\newline \textit{Model}: The encoder $\theta_f$ contains four $3 \times 3$ convolutional layers followed by a dense layer. The convolutional layers contain 64, 128, 192, and 256 filters, respectively. All layers in $\theta_f$ are ReLU activated. The encoder $\theta_f$ accepts an input image size of $96 \times 96$, and the dimension of the output representation is $256$. The classifier head $\theta_c$ is a single dense layer with $10$ output neurons and softmax activation. A complete classification model $\theta$ can be obtained by directly joining $\theta_f$ and $\theta_c$ together.
\vspace{0.6em}
\newline \textit{Contrastive pre-training}: To generate positive samples for contrastive pre-training, we rely on a combination of data augmentations, including horizontal flip, random cropping, and color transformations including jitters, brightness, and channel manipulations \citep{chen2020simple}.
We train $\theta_f$ for $300$ epochs with a batch size of $500$ on an unlabeled dataset using the Adam optimizer and an initial learning rate of $0.001$.
SimCLR~\citep{chen2020simple}, BYOL~\citep{grill2020bootstrap}, and SimSiam~\citep{chen2021exploring} incorporate different architectural configurations for training; therefore, we lightly tune the hyperparameters for each method. Specifically, for the projection head $\theta_g$, we tune the number of neurons in the dense layers, setting it to either 256 or 512.
We pre-train multiple encoders with different hyperparameters and then select the best one through linear evaluation \citep{chen2020simple, grill2020bootstrap}.
Linear evaluation involves training and testing a classifier head on top of the frozen encoder using a proxy dataset, and the proxy test accuracy can be used to select the best encoder, which we transfer to the downstream task.
(We mainly use the STL-10 unlabeled portion for contrastive pre-training in our experiments and simply use the STL-10 training and testing portions as the proxy dataset for linear evaluation. It is worth mentioning again that the encoder is trained only on the unlabeled portion.)
Our pre-training configurations are described below:

\begin{itemize}
    \item \textit{SimCLR}: The projection head $\theta_g$ contains two dense layers with $256$ neurons each. ReLU activation is applied to the first layer. We set the value of temperature $\tau$ as $0.1$.

    \item \textit{BYOL}: Each of the two dense layers in $\theta_g$ contains $512$ neurons. Both layers are batch normalized, although only the first layer is ReLU activated. The predictor $\theta_q$ contains a bottleneck dense layer with $64$ neurons followed by an output dense layer with $512$ neurons. The bottleneck layer is batch normalized and ReLU activated. We set the target branch update rate $\beta$ as $0.9$.

    \item \textit{SimSiam}: The projection head $\theta_g$ and the predictor $\theta_q$ are configured similarly to the settings in BYOL.
\end{itemize}
\vspace{0.6em}
\textit{Clustered federated learning}: For the federated setting, we construct the local data of $60$ clients by sampling from the training portion of a labeled dataset. To obtain a clear cluster structure, the clients are divided into three groups with different underlying data distributions. An example with the STL-10 training portion is shown in Figure~\ref{fig:client_data_amount}. Each client contains data from four different classes, and clients within the same cluster share the same classes. Specifically, for a dataset with $10$ available classes, a group of clients may contain classes $1$--$4$, another group may contain classes $4$--$7$, and the third group may contain classes $7$--$10$. Since each client contains four classes, we randomly allocate $20$ images for two of the classes and five images for the remaining two (i.e., a total of 50 images per client) to reflect the non-IID settings of FL. We also construct local testing sets that follow each client's data distribution by sampling from the testing portion of the dataset. We set the number of clusters $N$ to $3$, the number of global rounds $T$ to $100$, and the number of classifier head training rounds $T_c$ to $10$. Each client trains its selected model for three local epochs using a batch size of $32$ and the Adam optimizer with a learning rate of $0.001$.

\begin{table}[b]
\centering
\begin{tabular}{l|lV{3}cccc}
\hlineB{3}
& {\multirow{2}{*}{\begin{tabular}[c]{@{}c@{}}Pre-\\ training\end{tabular}}} & \multicolumn{4}{c}{Accuracy (\%)}  \\
                                    &                   & $T=25$             & $T=50$             & $T=75$             & $T=100$                        \\ \hlineB{3}
\multirow{2}{*}{FedAvg}    & None     & 32.73          & 39.83          & 41.20          & 44.13                      \\
                                    & SimCLR   & 46.40          & 53.33          & 53.97          & 59.33                      \\ \hline
\multirow{2}{*}{IFCA}      & None     & 57.60          & 63.30          & 65.60          & 67.80                      \\
                                    & FedAvg   & 69.93          & 71.77          & 70.40          & 70.90                      \\ \hline
\multirow{3}{*}{CP-CFL}    & SimCLR   & \textbf{72.30} & \textbf{75.80} & \textbf{76.30} & \textbf{76.80}             \\
                                    & BYOL    & 67.37          & 72.83          & 74.03          & 74.87                      \\
                                    & SimSiam  & 56.90          & 64.23          & 66.80          & 68.23                      \\ \hlineB{3}
\end{tabular}
\caption{Test accuracy $(\%)$ for the STL-10-to-STL-10 (denoting pre-training data to client data) task. The results are obtained by averaging the local test accuracies of all the clients.}
\label{tab:different_methods}
\end{table}

\begin{table}[htbp]
\centering
\resizebox{\linewidth}{!}{%

\begin{tabular}{l|lV{3}cc|cccc}
\hlineB{3}
                                 & \multicolumn{1}{cV{3}}{\multirow{3}{*}{\begin{tabular}[c]{@{}c@{}}Pre-\\ training\end{tabular}}} & \multicolumn{2}{c|}{F1 score} & \multicolumn{4}{c}{AUROC}                                                                                                                                                                                                                      \\ \cline{3-8}
\textbf{}                        & \multicolumn{1}{cV{3}}{}                                                                                  & Macro   & Weighted   & \begin{tabular}[c]{@{}c@{}}OvR\\ Macro\end{tabular} & \begin{tabular}[c]{@{}c@{}}OvR\\ Weighted\end{tabular} & \begin{tabular}[c]{@{}c@{}}OvO\\ Macro\end{tabular} & \begin{tabular}[c]{@{}c@{}}OvO\\ Weighted\end{tabular} \\ \hlineB{3}
\multirow{2}{*}{FedAvg} & None                                                                                          & 0.3708           & 0.3953              & 0.8342                                                       & 0.8333                                                          & 0.8374                                                       & 0.8349                                                          \\
                                 & SimCLR                                                                                        & 0.5564           & 0.5699              & 0.9036                                                       & 0.9045                                                          & 0.9068                                                       & 0.9052                                                          \\ \hline
\multirow{2}{*}{IFCA}   & None                                                                                          & 0.6842           & 0.6769              & 0.9562                                                       & 0.9491                                                          & 0.9595                                                       & 0.9542                                                          \\
                                 & FedAvg                                                                                        & 0.7131           & 0.7075              & 0.9641                                                       & 0.9584                                                          & 0.9669                                                       & 0.9626                                                          \\ \hline
\multirow{3}{*}{CP-CFL} & SimCLR                                                                                        & \textbf{0.7700}           & \textbf{0.7657}              & \textbf{0.9752}                & \textbf{0.9717}                         & \textbf{0.9774}                 & \textbf{0.9745} \\
                                 & BYOL                                                                                          & 0.7494           & 0.7455              & 0.9733                                                       & 0.9692                                                          & 0.9755                                                       & 0.9722                                                          \\
                                 & SimSiam                                                                                       & 0.6870           & 0.6827              & 0.9578                                                       & 0.9511                                                          & 0.9609                                                       & 0.9559                                                          \\ \hlineB{3}
\end{tabular}}

\caption{F1 score and AUROC for the STL-10-to-STL-10 task ($T=100$). OvR, one-vs-rest. OvO, one-vs-one.}
\label{tab:addition_scores}
\end{table}

\begin{figure}[htb]
    \centering
    \includegraphics[width=0.8\textwidth]{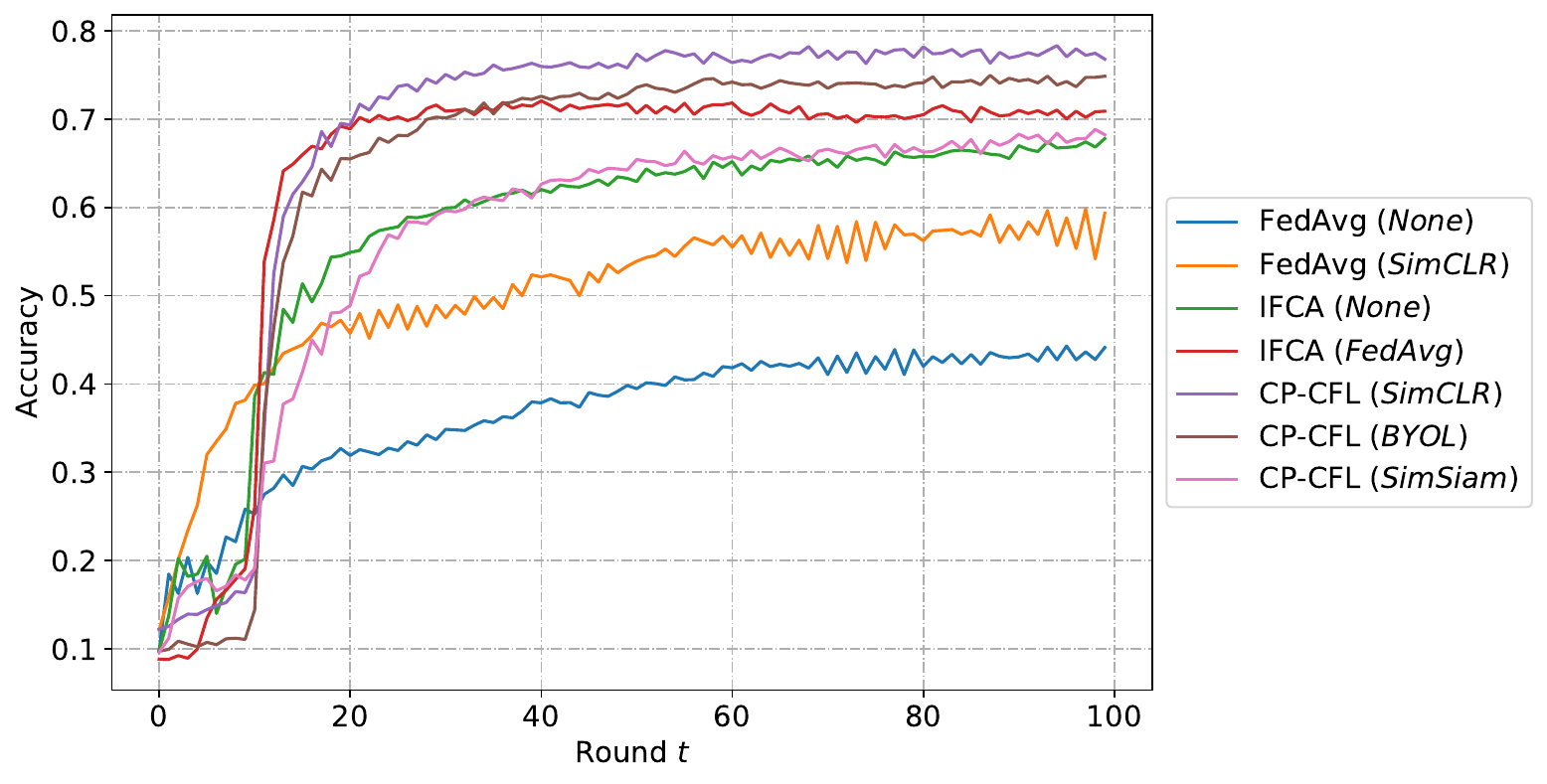}
    \caption{Test accuracy curves for the STL-10-to-STL-10 task.}
    \label{fig:table2_accuracy}
\end{figure}

\subsection{Encoder pre-training with contrastive learning}
\label{sec:pre_train_encoders}

To validate the effectiveness of contrastive pre-training for the downstream CFL task, we pre-train the encoder using SimCLR~\citep{chen2020simple}, BYOL~\citep{grill2020bootstrap}, and SimSiam~\citep{chen2021exploring}. We use the STL-10 unlabeled portion for pre-training, while the labeled training and testing portions are used to set up the clients' data as mentioned in Section~\ref{sec:exp_settings}. Figure~\ref{fig:client_data_amount} illustrates the distribution of private data across the clients. Although the data distribution of the STL-10 unlabeled portion is similar to that of the labeled portion, it is not identical. This appropriately reflects the practical scenarios in which the central server may not be able to gather pre-training data that exactly match the clients' data distribution.

Table~\ref{tab:different_methods} compares the performance of CP-CFL with that of different baselines, including vanilla FedAvg~\citep{mcmahan2017communication}, and IFCA~\citep{ghosh2020efficient} with and without a pre-trained encoder.
We denote the pre-training approaches in italics in parentheses after the abbreviations for the different methods.
IFCA (\textit{FedAvg}) uses the encoder obtained from the FedAvg training as the pre-trained encoder.
(Using the encoder from the FL global model is one option for getting a pre-trained encoder for the CFL task. However, this can be considered as pre-training on the clients, which would impose additional communication and computational burden on the clients.)
In the event that clustering fails for IFCA, we restart the process with a different set of randomly initialized parameters.
For the evaluation results, we average the local test accuracies from all clients, obtained by their respective cluster models.
From Table~\ref{tab:different_methods} and Figure~\ref{fig:table2_accuracy}, we can observe that \text{CP-CFL (\textit{SimCLR})} and \text{CP-CFL (\textit{BYOL})} significantly outperform all the baselines at $T=100$.
\text{CP-CFL (\textit{SimSiam})} outperforms IFCA~(\textit{None}) by a small margin but performs worse than IFCA (\textit{FedAvg}). However, recall that IFCA (\textit{FedAvg}) uses the encoder directly trained on the clients' labeled data, whereas SimCLR, BYOL, and SimSiam pre-train on unlabeled data.
The inefficiency of vanilla FedAvg in aggregating heterogeneous parameters degrades the resulting encoder quality, leading to lower downstream task performance for IFCA (\textit{FedAvg}) compared with CP-CFL~(\textit{SimCLR}) and CP-CFL (\textit{BYOL}).
Despite having a higher communication cost, CP-CFL can significantly outperform FedAvg even with a small number of global rounds $T$, as shown in Figure~\ref{fig:table2_accuracy}.

\begin{figure}[htb]
    \centering
    \includegraphics[width=0.8\textwidth]{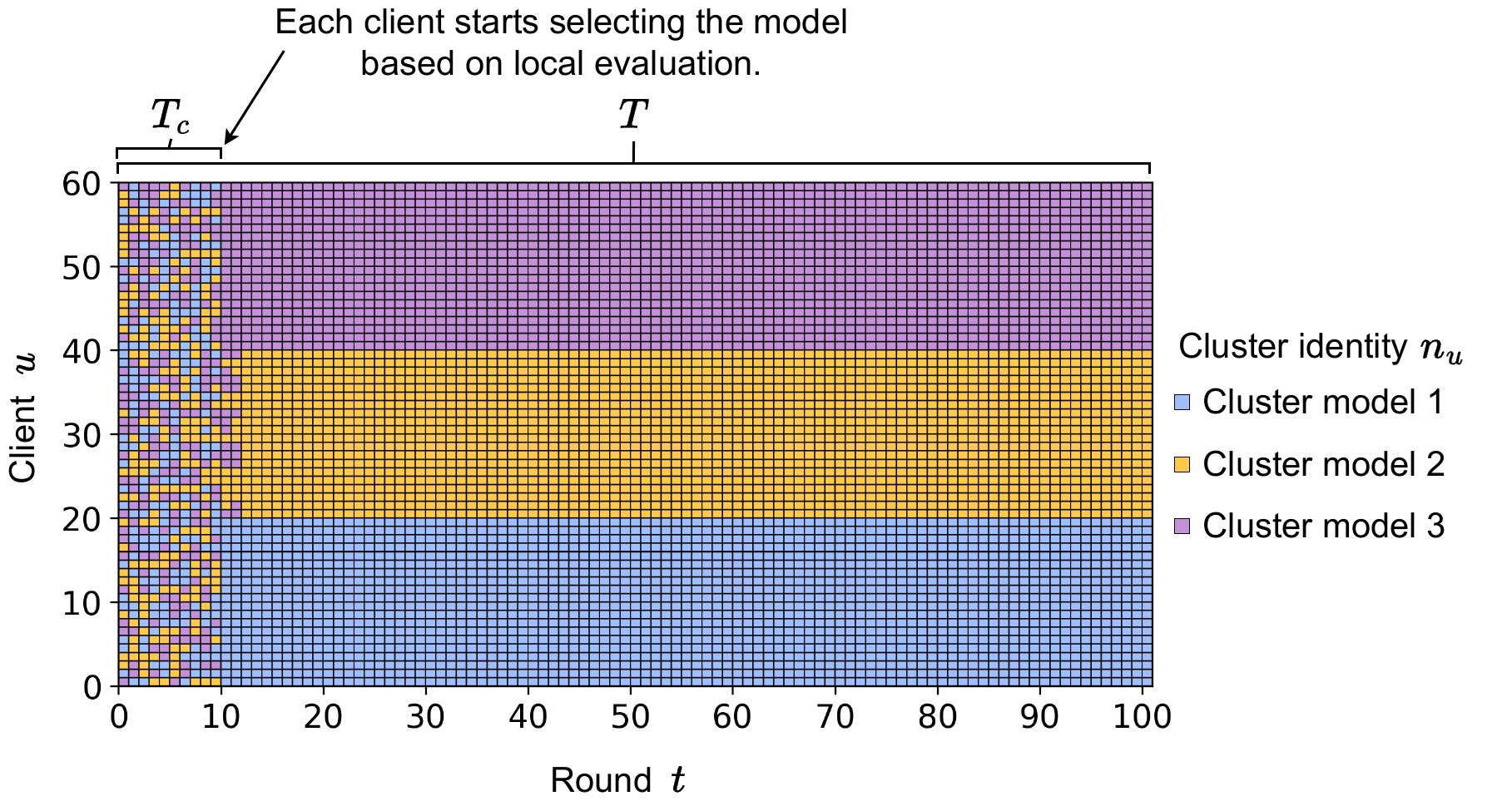}
    \caption{Cluster identity of clients at each communication round for CP-CFL~(\textit{SimCLR}).}
    \label{fig:cluster_identity}
\end{figure}

We report additional evaluation metrics---i.e., F1 score and area under the receiver operating characteristic (AUROC)---in Table~\ref{tab:addition_scores}. 
Figure~\ref{fig:cluster_identity} shows how the cluster identity $n_u$ of each client $u$ changes at different rounds for \text{CP-CFL}~(\textit{SimCLR}). From Figure~\ref{fig:cluster_identity}, we can observe that correct client clusters are identified almost immediately when the clients start selecting models.


\begin{table}[tb]
\centering
\begin{tabular}{l|lV{3}cccc}
\hlineB{3}
                            &       & \multicolumn{4}{c}{Pre-training data to client data}                    \\ \cline{3-6}
                            & \begin{tabular}[c]{@{}c@{}}Pre-\\ training\end{tabular} & \begin{tabular}[c]{@{}c@{}}STL-10\\ to \\ STL-10\\ (0.7346)\end{tabular}   & \begin{tabular}[c]{@{}c@{}}STL-10\\ to\\ CIFAR-10\\ (0.7306)\end{tabular}    & \begin{tabular}[c]{@{}c@{}}STL-10\\ to \\ MNIST\\ (0.6108)\end{tabular}   & \begin{tabular}[c]{@{}c@{}}EMNIST\\ to \\ MNIST\\ (0.6714)\end{tabular} \\ \hlineB{3}
\multirow{2}{*}{FedAvg}     & None          & 44.13               & 40.10                 & 95.47             & 95.47               \\
                            & SimCLR        & 59.33               & 48.20                 & 93.03             & 94.83               \\ \hline
\multirow{2}{*}{IFCA}        & None          & 67.80               & 65.40                 & 98.37             & 98.37               \\
                            & FedAvg        & 70.90               & 70.03                 & \textbf{98.90}    & 98.90               \\ \hline
CP-CFL                      & SimCLR        & \textbf{76.80}      & \textbf{75.00}        & 98.40             & \textbf{99.27}      \\ \hlineB{3}

Centralized (all clients)   & SimCLR        & 64.37               & 63.03                 & --             & --                   \\
Centralized (by cluster)    & SimCLR        & 77.47               & 76.00                 & --             & --                   \\ \hlineB{3}
\end{tabular}
\caption{Test accuracy $(\%)$ of CP-CFL (\textit{SimCLR}) with different pre-training and downstream client datasets at $T=100$. Note that pre-training data are unlabeled. The measure of the pre-training dataset's relevance to the downstream client dataset is shown in parentheses for each task.}
\label{tab:different_datasets}
\end{table}

\subsection{Influence of pre-training and downstream client data similarity }
\label{sec:sim_pre_downstream_data}

In practice, the pre-training data collected by the central server may not perfectly match the private data generated by the clients.
The distribution shift and differences in characteristics between the pre-training and downstream datasets could influence the model performance.
To investigate this, we pre-train the encoder using SimCLR on the STL-10 unlabeled portion and then transfer it to three different downstream CFL tasks, where clients' private and testing data are sampled from the STL-10, CIFAR-10~\citep{krizhevsky2009learning}, and MNIST~\citep{lecun1998gradient} datasets, respectively.
We also pre-train the encoder using the EMNIST dataset~\citep{cohen2017emnist}  without using labels and then transfer it to the MNIST task. Table~\ref{tab:different_datasets} shows the evaluation results as well as a measure of the pre-training dataset's relevance to the downstream client data in parentheses.

To quantify the relevance between the pre-training and downstream datasets for each task, we randomly sample 300 images from each dataset and extract their representations using the corresponding pre-trained encoder. We then measure their cross similarities using the cosine similarity, Eq.~\eqref{eqn:cosine_similarity}, and report the average score.
The results, shown in Table~\ref{tab:different_datasets}, indicate that the pre-training data and downstream client data share relevant characteristics for STL-10-to-STL-10, STL-10-to-CIFAR-10, and EMNIST-to-MNIST tasks, as evidenced by their relatively higher measurement scores compared with the score for the STL-10-to-MNIST task, where the pre-training data and downstream client data do not share relevant characteristics.

Table~\ref{tab:different_datasets} confirms that contrastive pre-training can significantly improve the downstream CFL performance if the unlabeled pre-training data share relevant characteristics with the downstream client data.
This is evident in STL-10-to-STL-10, STL-10-to-CIFAR-10, and EMNIST-to-MNIST tasks, where CP-CFL~(\textit{SimCLR}) outperforms IFCA~(\textit{FedAvg}) by $5.90\%$, $4.97\%$, and $0.37\%$, respectively.
In contrast, when the STL-10 pre-trained encoder is transferred to MNIST client data, there is no improvement over the baseline IFCA~(\textit{FedAvg}) (decreases by $0.5\%$). It is worth noting that the unlabeled portion of the STL-10 dataset contains extra image classes that are not in the labeled portion.
In addition, the CIFAR-10 dataset includes the ``frog'' class, as opposed to the ``monkey'' class present in the STL-10 dataset.
The significant performance gains observed in STL-10-to-STL-10 and STL-10-to-CIFAR-10 tasks suggest that self-supervised contrastive pre-training can be beneficial even if the pre-training data are not identical to the downstream data. However, it is ineffective when  the encoder is transferred between vastly dissimilar data, such as in the STL-10-to-MNIST task.

Table~\ref{tab:different_datasets} also presents the evaluation results for two centralized settings.
In the first setting, we gather and consolidate the private data of all clients into a single dataset and train a model in the centralized manner.
We evaluate the model on the testing dataset created in the same manner.
In the second setting, we gather and group the clients' data into three separate datasets based on their true underlying cluster structures; therefore, this setting comprises three pairs of training and testing datasets.
We train and evaluate a model for each pair and report the average results.
Notably, the performance of CP-CFL is comparable to that of the second centralized setting, which can be regarded as the upper bound.

\begin{figure}[b]
    \centering
    \includegraphics[width=0.5\textwidth]{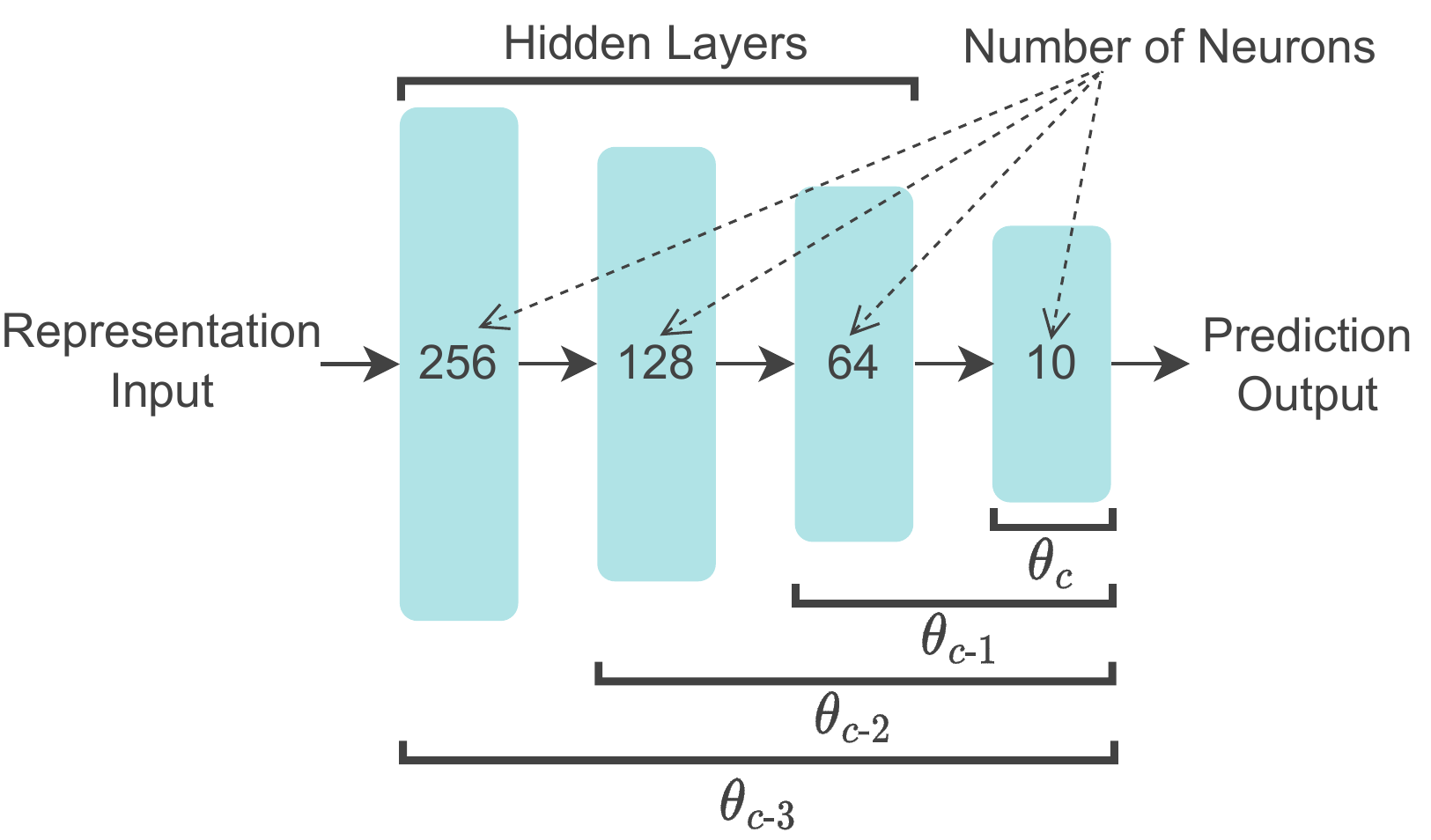}
    \caption{Detailed structure of classifier heads: $\theta_{c}$, $\theta_{c\text{-}1}$, $\theta_{c\text{-}2}$, and $\theta_{c\text{-}3}$. The default classifier head $\theta_c$  contains an output layer with 10 neurons. $\theta_{c\text{-}1}$ contains one more dense layer, in addition to the output layer. $\theta_{c\text{-}2}$ contains two more dense layers, and so on.}
    \label{fig:head_structure}
\end{figure}

\subsection{Learning capacity of the classifier head}
\label{sec:classifier_head}

The classifier head $\theta_c$ plays an important role in the downstream CFL performance as it primarily generates final class probabilities from the extracted representations. The learning capacity of $\theta_c$ can significantly affect the performance of a classification model $\theta=(\theta_f, \theta_c)$. To investigate this, we study four different classifier heads: $\theta_{c}$, $\theta_{c\text{-}1}$, $\theta_{c\text{-}2}$, and $\theta_{c\text{-}3}$, which differ in the number of layers and neurons. Here $\theta_{c}$ is our default classifier head, which contains a single output layer. The other classifier heads, $\theta_{c\text{-}1}$, $\theta_{c\text{-}2}$, and $\theta_{c\text{-}3}$, contain one, two, and three additional dense layers with ReLU activation, respectively. The details of different classifier head structures are illustrated in Figure~\ref{fig:head_structure}.

Table~\ref{tab:head_different_dataset} shows the performance of CP-CFL~(\textit{SimCLR}) with different classifier heads. Compared with $\theta_{c}$, having extra hidden layers in the classifier head (i.e., $\theta_{c\text{-}1}$, $\theta_{c\text{-}2}$, and $\theta_{c\text{-}3}$) can result in better performance when the number of global rounds is limited. For instance, in the STL-10-to-STL-10 task, $\theta_{c\text{-}3}$ at $T=25$ achieves similar performance as $\theta_{c}$ at $T=100$. Such an observation can be particularly useful for cases when client devices have limited communication resources, and we can trade off the computational cost of a few extra layers to obtain an acceptable performance with fewer global rounds. The performance gap between $\theta_c$ and other, larger classifier heads slowly diminishes as the number of rounds increases. (The results for the STL-10-to-MNIST task are provided in \ref{sec:app_classifier_head}.)

\begin{table}[t]
\centering
\resizebox{\linewidth}{!}{%
\begin{tabular}{lV{3}ccc|ccc|ccc}
\hlineB{3}
\multicolumn{1}{cV{3}}{}    & \multicolumn{3}{c|}{\begin{tabular}[c]{@{}c@{}}STL-10-to-STL-10\end{tabular}} & \multicolumn{3}{c|}{\begin{tabular}[c]{@{}c@{}}STL-10-to-CIFAR-10\end{tabular}} & \multicolumn{3}{c}{\begin{tabular}[c]{@{}c@{}}EMNIST-to-MNIST\end{tabular}} \\
\multicolumn{1}{cV{3}}{} & $T=25$                        & $T=75$                        & $T=100$                       & $T=25$                         & $T=75$                         & $T=100$                       & $T=25$                        & $T=75$                       & $T=100$ \\ \hlineB{3}
$\theta_{c}$             & 72.30                     & 76.30                     & 76.80                     & 67.20                      & 74.63                      & 75.00                     & 96.20                     & \textbf{99.20}           & \textbf{99.27}           \\
$\theta_{c\text{-}1}$    & 74.47                     & 77.77                     & \textbf{78.23}            & \textbf{70.23}             & 74.40                      & \textbf{75.70}            & 96.87                     & 98.87                    & 98.93                    \\
$\theta_{c\text{-}2}$    & 75.43                     & \textbf{78.57}            & 77.60                     & 69.77                      & 74.57                      & 75.10                     & 97.67                     & 99.03                    & 98.93                    \\
$\theta_{c\text{-}3}$    & \textbf{76.37}            & 78.43                     & 77.90                     & 70.00                      & \textbf{74.83} & 74.67                     & \textbf{97.73}            & 98.90                    & 98.83                    \\ \hlineB{3}
\end{tabular}
}%
\caption{Test accuracy $(\%)$ of CP-CFL~(\textit{SimCLR}) with different classifier head structures.}
\label{tab:head_different_dataset}
\end{table}

\begin{table}[b]
\centering
\begin{tabular}{lV{3}cccc}
\hlineB{3}
\multicolumn{1}{cV{3}}{\multirow{2}{*}{\begin{tabular}[c]{@{}c@{}} Local training \\ frequency of $\theta_f$\end{tabular}}} & \multicolumn{4}{c}{Classifier head structure}                                         \\
\multicolumn{1}{cV{3}}{}                                                                                              & $\theta_{c}$ & $\theta_{c\text{-}1}$ & $\theta_{c\text{-}2}$ & $\theta_{c\text{-}3}$ \\ \hlineB{3}
$E_{\ell,f} = 3$ (default)                                                                   & 76.80                 & \textbf{78.23}                 & \textbf{77.60}        & \textbf{77.90}                 \\
$E_{\ell,f} = 1$                                                                                      & \textbf{76.90}        & 74.93                 & 74.80                 & 73.97                 \\
$E_{\ell,f} = 0$ (frozen)                                                                    & 73.67                 & 77.17                 & 77.57                 & 76.77                 \\ \hlineB{3}
$E_{\ell,f} = 3$ (global)                                                                    & 76.27                 & 79.10        & 77.50                 & 78.03        \\ \hlineB{3}
\end{tabular}
\caption{Test accuracy $(\%)$ of CP-CFL~(\textit{SimCLR}) for the STL-10-to-STL-10 task ($T=100$). $E_{\ell,f}$ is the number of local epochs for training the encoder $\theta_f$.}
\label{tab:encoder_training}
\end{table}

\subsection{Encoder in the local training step}
\label{sec:encoder_training}

In CP-CFL, clients train the entire model (both the encoder $\theta_f$ and the classifier head $\theta_c$) at each global round $t>T_c$. However, it might be possible to freeze the pre-trained encoder $\theta_f$ in all training rounds without a significant impact on downstream CFL performance. Since $\theta_f$ has already been pre-trained on the unlabeled data, we can lighten the computational load on the clients by reducing the frequency of local training for $\theta_f$. Therefore, we study three different settings in which the encoder $\theta_f$ is trained for zero, one, and three (default) local epochs. The classifier head $\theta_c$ is still trained for three local epochs in all settings. The experiments are conducted on the STL-10-to-STL-10 task with different classifier head structures.

From the results in Table~\ref{tab:encoder_training}, it is clear that training the encoder $\theta_f$ for three local epochs ($E_{\ell,f}=3$) generally achieves the best performance or performance comparable to that of other settings. When used together with large classifier heads such as $\theta_{c\text{-}1}$, $\theta_{c\text{-}2}$, and $\theta_{c\text{-}3}$, freezing the encoder ($E_{\ell,f}=0$) still performs comparably to $E_{\ell,f}=3$ (with only minor accuracy drops of $1.06\%$, $0.03\%$, and $1.13\%$, respectively). This suggests that it may be possible to reduce the computational cost on the clients by freezing the encoder while using a large classifier head, with minimal accuracy reduction. (It is worth noting that the number of trainable parameters in a large classifier head, $\theta_{c\text{-}3} \approx 0.1$ M, is still significantly lower than that of the entire CNN encoder, $\theta_f \approx 2.4$ M.) Interestingly, training the encoder for a single local epoch ($E_{\ell,f}=1$) results in worse performance than the frozen encoder ($E_{\ell,f}=0$) for $\theta_{c\text{-}1}$, $\theta_{c\text{-}2}$, and $\theta_{c\text{-}3}$. We believe that inadequate training disrupts the pre-trained parameters in the encoder, hurting the compatibility between the encoder and the classifier head.


Furthermore, we study the global encoder setting, where all encoders from different client clusters are aggregated into a single global encoder. In this setting, all cluster models share the common global encoder, and only the classifier head is cluster specific. As shown in Table~\ref{tab:encoder_training}, the global encoder setting achieves promising results and presents an alternative to conventional CFL techniques where cluster models do not share the parameters.

\subsection{Supervised versus contrastive pre-training}
\label{sec:supervised_vs_contrastive}

CP-CFL pre-trains the encoder $\theta_f$ on unlabeled data using contrastive learning. It is crucial to determine the potential performance drop caused by use of unlabeled data instead of labeled data for pre-training. To this end, we pre-train $\theta_f$ using both supervised and contrastive settings and compare their performance in the downstream CFL task. We use the CIFAR-100  dataset~\citep{krizhevsky2009learning} for centralized pre-training and the CIFAR-10 dataset to construct the private data of clients. In the supervised pre-training setting, a classification model $\theta=(\theta_f, \theta_c)$ is trained with cross-entropy loss on the CIFAR-100 dataset, and then we keep only the encoder $\theta_f$. In the contrastive learning setting, we use SimCLR to pre-train the encoder on the CIFAR-100 dataset without using the labels.

Table~\ref{tab:encoder_pretraining} compares the downstream CFL performance between different encoder pre-training settings. Supervised pre-training outperforms contrastive pre-training when $T$ is small (i.e., $T \approx 25$). However, with a higher number of global rounds (i.e., $T \geq 50$), the performance gap between the two pre-training methods becomes negligible. Therefore, we can conclude that the encoder pre-trained with contrastive learning performs effectively in the downstream CFL task.

\begin{table}[t]
\centering
\begin{tabular}{lV{3}cccc}
\hlineB{3}

\multirow{2}{*}{Pre-training} & \multicolumn{4}{c}{Accuracy (\%)} \\

   & $T=25$            & $T=50$            & $T=75$            & $T=100$               \\ \hlineB{3}
Contrastive        & 70.13         & 76.50         & \textbf{77.17} & \textbf{77.33}   \\
Supervised          & \textbf{72.83} & \textbf{76.80} & 77.13         & 77.13             \\ \hlineB{3}
\end{tabular}
\caption{Test accuracy (\%) for the CIFAR-100-to-CIFAR-10 task using supervised and contrastive encoder pre-training settings.}
\label{tab:encoder_pretraining}
\end{table}

\subsection{Number of clusters}
\label{sec:number_of_clusters}

CP-CFL does not explicitly compute the number of clusters $N$, since it is primarily determined by available system resources in practice. Although we set the number of clusters $N$ to $3$ by default in our experiments, it would be interesting to see how $N$ affects the performance of CP-CFL. Therefore, we conduct experiments by adjusting the value of $N$ between 1 and 5 on the STL-10-to-STL-10 task.

The results of CP-CFL (\textit{SimCLR}) with different numbers of clusters are reported in Table~\ref{tab:different_clusters}. Setting the number of clusters $N$ to $3$ gives the highest performance, since it matches the underlying cluster structure. However, use of a higher number of clusters, such as four or five, still yields better results than a lower number of clusters, such as one or two. In general, increasing the number of clusters $N$ means improving the personalized performance at the expense of more system resources. Therefore, we can adjust $N$ to strike a balance between performance gain and resource expense.

\begin{table}[t]
\centering
\begin{tabular}{cV{3}c}
\hlineB{3}
\begin{tabular}[c]{@{}c@{}}Number of\\clusters ($N$)\end{tabular} & Accuracy (\%)         \\ \hlineB{3}
1         & 59.33            \\
2                  & 65.10            \\
3        & \textbf{76.80}   \\
4                  & 72.80            \\
5                  & 72.97            \\ \hlineB{3}
\end{tabular}
\caption{Test accuracy (\%) of CP-CFL~(\textit{SimCLR}) with different numbers of clusters for the STL-10-to-STL-10 task ($T=100$). The best result is shown in bold.}
\label{tab:different_clusters}
\end{table}

\subsection{Number of clients}

In this section, we investigate the impact of the number of clients on CP-CFL using the STL-10-to-STL-10 task. Specifically, we consider four different settings, where the number of clients is set to 15, 30, 60 (default), and 90, respectively. In each setting, the clients are equally divided into three groups with different underlying data
distributions, as described in Section~\ref{sec:exp_settings}. We use the default hyperparameters, with the exception of setting $T=40$ and $T=60$ for the 15-client and 30-client settings, respectively.

\begin{figure}[htb]
    \centering
    \subfloat[15 clients]{\includegraphics[width=0.4\textwidth]{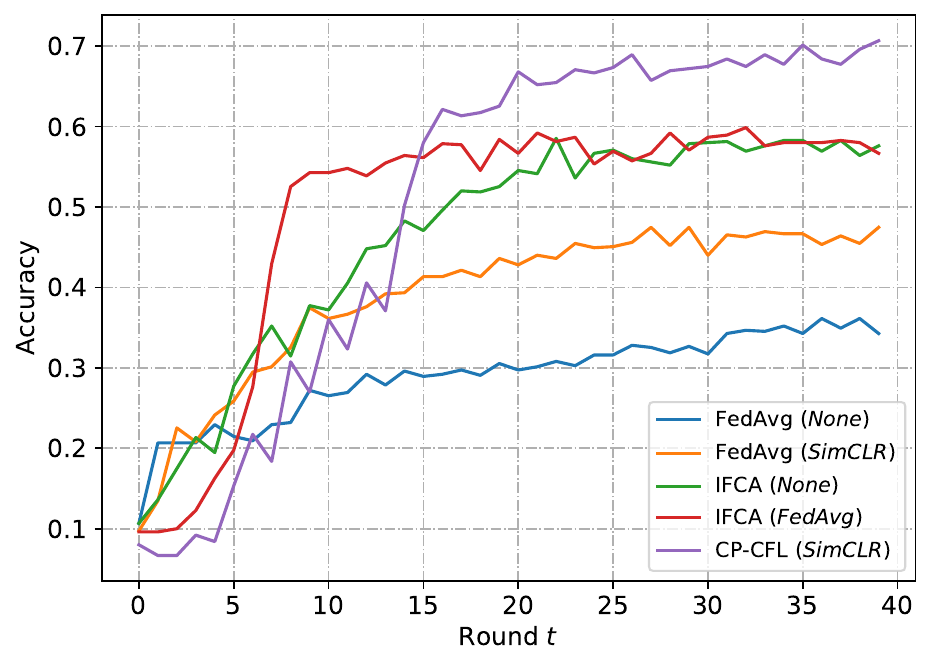}
     \label{subfig:c_15}}
	\subfloat[30 clients]{\includegraphics[width=0.4\textwidth]{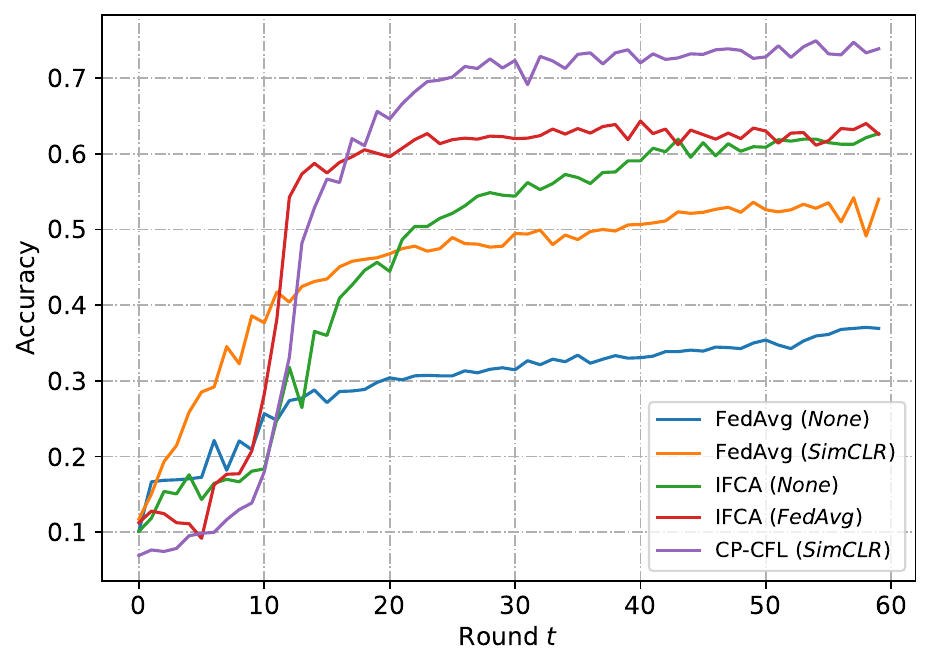}
 \label{subfig:c_30}} \\
	\subfloat[60 clients]{\includegraphics[width=0.4\textwidth]{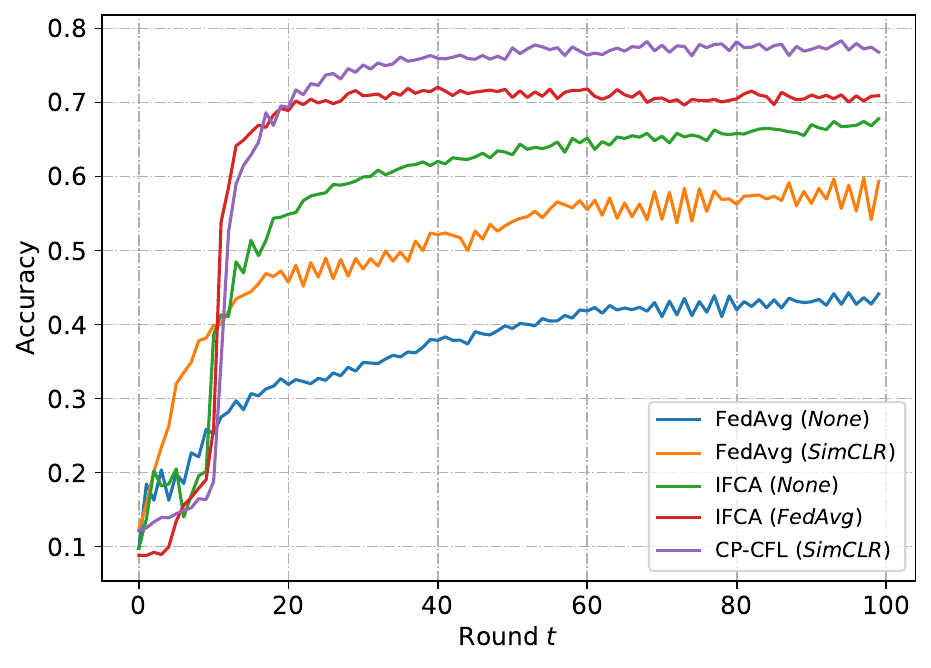}
 \label{subfig:c_60}}
    \subfloat[90 clients]{\includegraphics[width=0.4\textwidth]{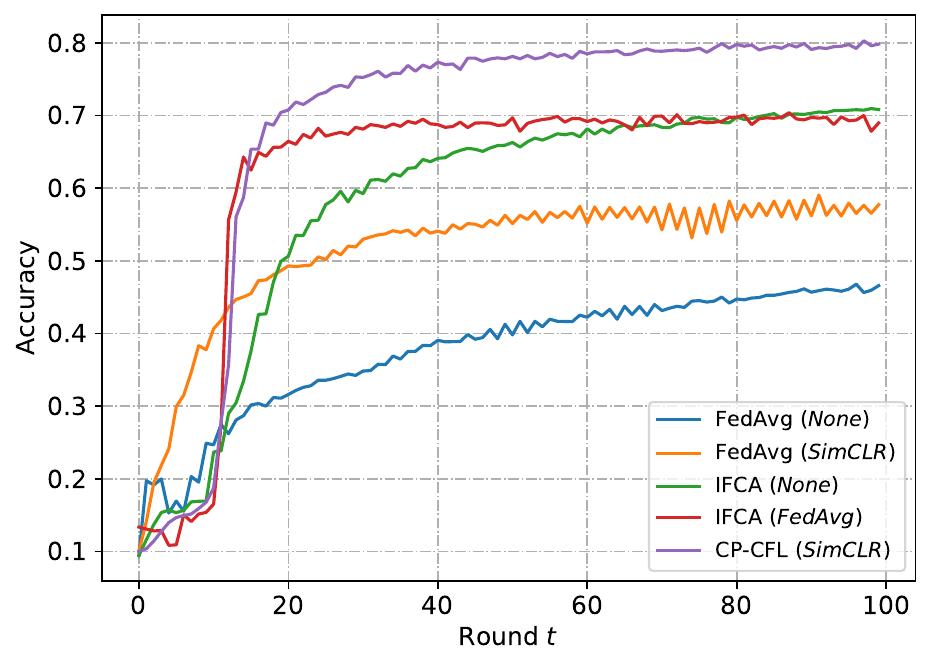}
     \label{subfig:c_90}}
	\caption{Test accuracy curves obtained with different numbers of clients on the STL-10-to-STL-10 task.}
	\label{fig:num_clients}
\end{figure}

Since the number of samples per client is fixed (i.e., 50), increasing the number of clients results in more training data. As presented in Table~\ref{tab:different_number_of_clients}, the performance of CP-CFL (\textit{SimCLR}) improves as the number of clients increases, a trend that is also observed in other approaches. While there is no significant difference in the final performance between IFCA~(\textit{None}) and IFCA~(\textit{FedAvg}) in Table~\ref{tab:different_number_of_clients}, Figure~\ref{fig:num_clients} reveals that IFCA~(\textit{FedAvg}) exhibits a performance boost in early training rounds. This trend holds for all approaches involving pre-training, i.e., FedAvg~(\textit{SimCLR}), IFCA~(\textit{FedAvg}), and CP-CFL~(\textit{SimCLR}).

\begin{table}[htb]
\centering
\resizebox{\linewidth}{!}{%

\begin{tabular}{l|lV{3}cccccccc}
\hlineB{3}
                & \multicolumn{1}{cV{3}}{\multirow{3}{*}{\begin{tabular}[c]{@{}c@{}}Pre-\\ training\end{tabular}}} & \multicolumn{8}{c}{Number of clients}                                                                                                     \\ \cline{3-10}
 \textbf{}       & \multicolumn{1}{cV{3}}{}                                                                                  & \multicolumn{2}{c|}{15}    & \multicolumn{2}{c|}{30}    & \multicolumn{2}{c|}{60}     & \multicolumn{2}{c}{90} \\ \cline{3-10}
                & \multicolumn{1}{cV{3}}{}                                                                                  & $T=40$ & \multicolumn{1}{c|}{Best}  & $T=60$ & \multicolumn{1}{c|}{Best}  & $T=100$ & \multicolumn{1}{c|}{Best}  & $T=100$         & Best          \\ \hlineB{3}
\multirow{2}{*}{FedAvg} & None                                                                                          & 34.27  & \multicolumn{1}{c|}{36.13} & 36.93  & \multicolumn{1}{c|}{37.07} & 44.13   & \multicolumn{1}{c|}{50.73} & 46.58           & 46.78         \\
\textbf{}       & SimCLR                                                                                        & 47.47  & \multicolumn{1}{c|}{47.47} & 54.00  & \multicolumn{1}{c|}{54.20} & 59.33   & \multicolumn{1}{c|}{59.83} & 57.71           & 59.04         \\ \hline
\multirow{2}{*}{IFCA}   & None                                                                                          & 57.60  & \multicolumn{1}{c|}{58.53} & 62.67  & \multicolumn{1}{c|}{62.67} & 67.80   & \multicolumn{1}{c|}{67.80} & 70.80           & 70.93         \\
\textbf{}       & FedAvg                                                                                        & 56.67  & \multicolumn{1}{c|}{59.87} & 62.60  & \multicolumn{1}{c|}{64.33} & 70.90   & \multicolumn{1}{c|}{72.07} & 68.95           & 70.36         \\ \hline
CP-CFL & SimCLR                                                                                        & \textbf{70.67}  & \multicolumn{1}{c|}{\textbf{70.67}} & \textbf{73.87}  & \multicolumn{1}{c|}{\textbf{74.93}} & \textbf{76.80}   & \multicolumn{1}{c|}{\textbf{78.33}} & \textbf{79.80}           & \textbf{80.27} \\ \hlineB{3}
\end{tabular}}

\caption{Test accuracy $(\%)$ of CP-CFL (\textit{SimCLR}) with different numbers of clients for the STL-10-to-STL-10 task. We report both the final results and the best results.}
\label{tab:different_number_of_clients}
\end{table}

\subsection{Statistical significance testing}
\label{sec:client_data_partition}

To ensure the statistical significance of our study, we conduct five independent experiments on the STL-10-to-STL-10 task. For each experiment, we resample the private and testing data of clients according to the settings described in Section~\ref{sec:exp_settings}. Table~\ref{tab:different_client_partitions} reports the evaluation results of all five experiments, as well as their mean and standard deviation. Our comparison with baseline approaches demonstrates that CP-CFL~(\textit{SimCLR}) maintains good results across all trials, and outperforms  IFCA~(\textit{None}) by $10.90\%$ and IFCA~(\textit{FedAvg}) by $8.35\%$ on average.
Additional metrics are provided in \ref{sec:app_additional_metrics}.

\begin{table}[htb]
\centering

\begin{tabular}{l|lV{3}ccccc|c}
\hlineB{3}
                                 & \multicolumn{1}{cV{3}}{\multirow{2}{*}{\begin{tabular}[c]{@{}c@{}}Pre-\\ training\end{tabular}}} & \multicolumn{5}{c|}{Accuracy (\%)}                                      &                                               \\
\textbf{}                        & \multicolumn{1}{cV{3}}{}                                                                                  & Trial 1 & Trial 2 & Trial 3 & Trial 4 & Trial 5 & Mean $\pm$ SD \\ \hlineB{3}
\multirow{2}{*}{FedAvg} & None                                                                                          & 42.73          & 41.53          & 43.67          & 43.03          & 42.67          & $42.73 \pm 0.69$                                \\
                                 & SimCLR                                                                                        & 58.30          & 52.60          & 56.10          & 55.40          & 57.50          & $55.98 \pm 1.97$                                \\ \hline
\multirow{2}{*}{IFCA}   & None                                                                                          & 66.17          & 64.37          & 66.43          & 66.17          & 65.80          & $65.79 \pm 0.74$                                \\
                                 & FedAvg                                                                                        & 69.57          & 66.53          & 69.30          & 68.20          & 69.43          & $68.61 \pm 1.14$                                \\ \hline
CP-CFL                  & SimCLR                                                                                        & \textbf{76.13}          & \textbf{76.47}          & \textbf{77.43}          & \textbf{76.40}          & \textbf{77.00}          & \bm{$76.69 \pm 0.47$}                                \\ \hlineB{3}
\end{tabular}

\caption{Test accuracy (\%) on multiple trials of the STL-10-to-STL-10 task ($T=100$). SD, standard deviation.}
\label{tab:different_client_partitions}
\end{table}

\subsection{ResNet-18 encoder}
\label{sec:resent18_encoder}

In addition to our default CNN encoder, we also evaluate the performance of CP-CFL using the ResNet-18 encoder~\citep{he2016deep} on the STL-10-to-STL-10 task.  This experiment aims to examine the generalization ability of CP-CFL with regrad to different encoder architectures other than our default CNN encoder, and not necessarily to obtain the best performance. Therefore, we use default values for most hyperparameters except for a few that we specify below.
The number of epochs for contrastive pre-training is set to 500 with a batch size of 250.
The output dimension of the ResNet-18 encoder is 512, and we use $\theta_{c\text{-}1}$ as the classifier head.
(We empirically found that while using the ResNet-18 encoder with our default classifier head $\theta_c$, which contains a single output layer, FedAvg (\textit{SimCLR}) often gets stuck in local optima, resulting in low performance; herefore, we use $\theta_{c\text{-}1}$ for all approaches.)
We conduct five independent experiments, and Table~\ref{tab:resnet18_encoder_best} reports the best results achieved by each approach. We report the best results in Table~\ref{tab:resnet18_encoder_best} since the performance of FedAvg baselines can fluctuate. Additional results for $T=100$ are provided in Table~\ref{tab:app_resnet18_encoder_t} in \ref{sec:app_resnet_18}. We also conduct a single experiment using the ResNet-50 encoder~\citep{he2016deep}; the results are shown in Table~\ref{tab:app_resnet50_encoder} in \ref{sec:app_resnet_50}. While the ResNet-50 encoder is commonly used for many image processing tasks, the ResNet-18 encoder can be more practical for FL environments, where  client devices often have limited computing resources for locally training the ResNet-50 encoder.

\begin{table}[b]
\centering

\begin{tabular}{l|lV{3}ccccc|c}
\hlineB{3}
\textbf{}                        & \multirow{2}{*}{\begin{tabular}[c]{@{}c@{}}Pre-\\ training\end{tabular}} & \multicolumn{5}{c|}{Accuracy (\%)}                                         &                                      \\
\textbf{}                        &                                                                                   & Trial 1    & Trial 2 & Trial 3    & Trial 4    & Trial 5    & Mean $\pm$ SD \\ \hlineB{3}
\multirow{2}{*}{FedAvg} & None                                                                     & 37.37          & 38.40          & 37.77          & 34.83          & 39.40          & $37.55 \pm 1.52$                     \\
                                 & SimCLR                                                                   & 45.77          & 39.77          & 44.97          & 39.33          & 39.70          & $41.91 \pm 2.84$                     \\ \hline
\multirow{2}{*}{IFCA}                    & None                                                                     & 64.10          & 64.67          & 64.80          & 65.53          & 63.93          & $64.61 \pm 0.57$                     \\
                        & FedAvg                                                                   & 65.03          & 65.27          & 64.87          & 64.00          & 65.40          & $64.91 \pm 0.49$                     \\ \hline
CP-CFL                  & SimCLR                                                                   & \textbf{67.90} & \textbf{68.57} & \textbf{68.30} & \textbf{68.90} & \textbf{66.53} & \bm{$68.04 \pm 0.82$}            \\ \hlineB{3}
\end{tabular}

\caption{Test accuracy (\%) obtained with the ResNet-18 encoder on the STL-10-to-STL-10 task (best). SD, standard deviation.}
\label{tab:resnet18_encoder_best}
\end{table}

\section{Discussion and future work}

The influence of the amount of unlabeled pre-training data on the performance of the encoder could be a topic of interest. In practice, it is often the responsibility of the service provider or the server to gather an adequate amount of unlabeled data for pre-training the encoder. The server needs to determine the quantity of pre-training data based on various factors, such as the type of service being provided and the size of the model deployed. The server may also keep a small set of proxy test data to evaluate the performance of the pre-trained encoder and determine if the collected unlabeled data are sufficient in terms of both quality and size. While a larger quantity of pre-training data would generally result in a higher-quality encoder, it would also require a longer training time and more computing resources. Therefore, we leave the investigation of the trilemma between the amount of pre-training data, the required training resources, and the resulting encoder quality to future studies.

\section{Conclusions}

FL is essential for incorporating edge-generated data into intelligent systems without violating privacy regulations. However, non-IID data at the edge pose a fundamental challenge that limits FL from achieving the same level of performance as centralized training. To address this issue, we proposed CP-CFL, which combines contrastive encoder pre-training and client clustering to alleviate the performance drop caused by non-IID data in a heterogeneous FL environment. We conducted extensive experiments on CP-CFL in various settings to demonstrate its superior performance over the baseline approaches. Furthermore, we explored efficient ways to deploy CP-CFL, while also providing various ablation studies to validate its effectiveness. Overall, our study contributes to a better understanding of how contrastive encoder pre-training and client clustering can jointly improve the performance of FL on non-IID data.








\appendix

\section{Learning capacity of the classifier head}
\label{sec:app_classifier_head}

The results for CP-CFL~(\textit{SimCLR}) with different classifier head structures on the STL-10-to-MNIST task are given in Table \ref{tab:head_different_dataset_stl10_mnist}.

\begin{table}[htbp]
\centering
\begin{tabular}{lV{3}ccc}
\hlineB{3}
 & \multicolumn{3}{c}{\begin{tabular}[c]{@{}c@{}}STL-10-to-MNIST\end{tabular}} \\
 & $T=25$                           & $T=75$                          & $T=100$                         \\  \hlineB{3}
$\theta_{c}$             & 93.70                        & 97.63                       & 98.40                       \\
$\theta_{c\text{-}1}$    & \textbf{94.20}               & 98.27                       & 98.13                       \\
$\theta_{c\text{-}2}$    & 93.53                        & \textbf{98.53}              & \textbf{98.63}              \\
$\theta_{c\text{-}3}$    & \textbf{94.20}               & 98.13                      & 98.20                       \\  \hlineB{3}
\end{tabular}
\caption{Test accuracy $(\%)$ of CP-CFL~(\textit{SimCLR}) with different classifier head structures on the STL-10-to-MNIST task.}
\label{tab:head_different_dataset_stl10_mnist}
\end{table}

\section{Additional metrics}
\label{sec:app_additional_metrics}

F1 score and AUROC for statistical significance testing reported in Section~\ref{sec:client_data_partition} are given in Tables \ref{tab:app_f1_macro}--\ref{tab:app_rocauc_ovoweighted}.

\begin{table}[htbp]
\centering
\resizebox{\linewidth}{!}{%

\begin{tabular}{l|lV{3}ccccc|c}
\hlineB{3}
                                 & \multicolumn{1}{cV{3}}{\multirow{2}{*}{\begin{tabular}[c]{@{}c@{}}Pre-\\ training\end{tabular}}} & \multicolumn{5}{c|}{F1 score (macro)}                                           &                                               \\
\textbf{}                        & \multicolumn{1}{cV{3}}{}                                                                                  & Trial 1 & Trial 2 & Trial 3 & Trial 4 & Trial 5 & Mean $\pm$ SD \\ \hlineB{3}
\multirow{2}{*}{FedAvg} & None                                                                                          & 0.3570         & 0.3451         & 0.3566         & 0.3315         & 0.3415         & $0.3463 \pm 0.0096$                             \\
                                 & SimCLR                                                                                        & 0.5484         & 0.4746         & 0.5177         & 0.5118         & 0.5304         & $0.5166 \pm 0.0245$                             \\ \hline
\multirow{2}{*}{IFCA}   & None                                                                                          & 0.6701         & 0.6479         & 0.6684         & 0.6685         & 0.6668         & $0.6643 \pm 0.0083$                             \\
                                 & FedAvg                                                                                        & 0.7002         & 0.6681         & 0.6984         & 0.6876         & 0.7006         & $0.6910 \pm 0.0124$                             \\ \hline
CP-CFL                  & SimCLR                                                                                        & \textbf{0.7640}         & \textbf{0.7593}         & \textbf{0.7757}         & \textbf{0.7630}         & \textbf{0.7698}         & \bm{$0.7664 \pm 0.0057$} \\ \hlineB{3}
\end{tabular}}

\caption{F1 score (macro) on multiple trials of the STL-10-to-STL-10 task ($T=100$). SD, standard deviation.}
\label{tab:app_f1_macro}
\end{table}

\begin{table}[htbp]
\centering
\resizebox{\linewidth}{!}{%

\begin{tabular}{l|lV{3}ccccc|c}
\hlineB{3}
                                 & \multicolumn{1}{cV{3}}{\multirow{2}{*}{\begin{tabular}[c]{@{}c@{}}Pre-\\ training\end{tabular}}} & \multicolumn{5}{c|}{F1 score (weighted)}                                           &                                               \\
\textbf{}                        & \multicolumn{1}{cV{3}}{}                                                                                  & Trial 1 & Trial 2 & Trial 3 & Trial 4 & Trial 5 & Mean $\pm$ SD \\ \hlineB{3}
\multirow{2}{*}{FedAvg} & None                                                                                          & 0.3817         & 0.3662         & 0.3859         & 0.3633         & 0.3692         & $0.3733 \pm 0.0089$                             \\
                                 & SimCLR                                                                                        & 0.5641         & 0.4933         & 0.5311         & 0.5306         & 0.5440         & $0.5326 \pm 0.0231$                             \\ \hline
\multirow{2}{*}{IFCA}   & None                                                                                          & 0.6606         & 0.6404         & 0.6638         & 0.6602         & 0.6577         & $0.6565 \pm 0.0083$                             \\
                                 & FedAvg                                                                                        & 0.6942         & 0.6624         & 0.6922         & 0.6807         & 0.6925         & $0.6844 \pm 0.0120$                             \\ \hline
CP-CFL                  & SimCLR                                                                                        & \textbf{0.7575}         & \textbf{0.7560}         & \textbf{0.7714}         & \textbf{0.7601}         & \textbf{0.7655}         & \bm{$0.7621 \pm 0.0056$}                             \\ \hlineB{3}
\end{tabular}}

\caption{F1 score (weighted) on multiple trials of the STL-10-to-STL-10 task ($T=100$). SD, standard deviation.}
\label{tab:app_f1_weighted}
\end{table}

\begin{table}[htbp]
\centering
\resizebox{\linewidth}{!}{%

\begin{tabular}{l|lV{3}ccccc|c}
\hlineB{3}
                                 & \multicolumn{1}{cV{3}}{\multirow{2}{*}{\begin{tabular}[c]{@{}c@{}}Pre-\\ training\end{tabular}}} & \multicolumn{5}{c|}{AUROC (OvR macro)}                                       &                                               \\
\textbf{}                        & \multicolumn{1}{cV{3}}{}                                                                                  & Trial 1  & Trial 2  & Trial 3  & Trial 4  & Trial 5  & Mean $\pm$ SD \\ \hlineB{3}
\multirow{2}{*}{FedAvg} & None                                                                                          & 0.8321          & 0.8278          & 0.8355          & 0.8379          & 0.8306          & $0.8328 \pm 0.0036$                             \\
                                 & SimCLR                                                                                        & 0.9033          & 0.8928          & 0.8990          & 0.9019          & 0.9080          & $0.9010 \pm 0.0050$                             \\ \hline
\multirow{2}{*}{IFCA}   & None                                                                                          & 0.9544          & 0.9550          & 0.9553          & 0.9573          & 0.9542          & $0.9552 \pm 0.0011$                             \\
                                 & FedAvg                                                                                        & 0.9614          & 0.9571          & 0.9592          & 0.9593          & 0.9597          & $0.9593 \pm 0.0014$                             \\ \hline
CP-CFL                  & SimCLR                                                                                        & \textbf{0.9756} & \textbf{0.9753} & \textbf{0.9766} & \textbf{0.9761} & \textbf{0.9759} & \bm{$0.9759 \pm 0.0004$}                    \\ \hlineB{3}
\end{tabular}}

\caption{AUROC (OvR macro) on multiple trials of the STL-10-to-STL-10 task ($T=100$). OvR, one-vs-rest. SD, standard deviation.}
\label{tab:app_rocauc_ovrmacro}
\end{table}

\begin{table}[htbp]
\centering
\resizebox{\linewidth}{!}{%

\begin{tabular}{l|lV{3}ccccc|c}
\hlineB{3}
                                 & \multicolumn{1}{cV{3}}{\multirow{2}{*}{\begin{tabular}[c]{@{}c@{}}Pre-\\ training\end{tabular}}} & \multicolumn{5}{c|}{AUROC (OvR weighted)}                                    &                                               \\
\textbf{}                        & \multicolumn{1}{cV{3}}{}                                                                                  & Trial 1  & Trial 2  & Trial 3  & Trial 4  & Trial 5  & Mean $\pm$ SD \\ \hlineB{3}
\multirow{2}{*}{FedAvg} & None                                                                                          & 0.8291          & 0.8269          & 0.8341          & 0.8366          & 0.8300          & $0.8313 \pm 0.0035$                             \\
                                 & SimCLR                                                                                        & 0.9040          & 0.8930          & 0.8987          & 0.9020          & 0.9079          & $0.9011 \pm 0.0050$                             \\ \hline
\multirow{2}{*}{IFCA}   & None                                                                                          & 0.9465          & 0.9472          & 0.9481          & 0.9501          & 0.9461          & $0.9476 \pm 0.0014$                             \\
                                 & FedAvg                                                                                        & 0.9549          & 0.9500          & 0.9521          & 0.9525          & 0.9526          & $0.9524 \pm 0.0015$                             \\ \hline
CP-CFL                  & SimCLR                                                                                        & \textbf{0.9718} & \textbf{0.9720} & \textbf{0.9734} & \textbf{0.9732} & \textbf{0.9723} & \bm{$0.9725 \pm 0.0006$}                    \\ \hlineB{3}
\end{tabular}}

\caption{AUROC (OvR weighted) on multiple trials of the STL-10-to-STL-10 task ($T=100$). OvR, one-vs-rest. SD, standard deviation.}
\label{tab:app_rocauc_ovrweighted}
\end{table}

\begin{table}[htbp]
\centering
\resizebox{\linewidth}{!}{%

\begin{tabular}{l|lV{3}ccccc|c}
\hlineB{3}
                                 & \multicolumn{1}{cV{3}}{\multirow{2}{*}{\begin{tabular}[c]{@{}c@{}}Pre-\\ training\end{tabular}}} & \multicolumn{5}{c|}{AUROC (OvO macro)}                                       &                                               \\
\textbf{}                        & \multicolumn{1}{cV{3}}{}                                                                                  & Trial 1  & Trial 2  & Trial 3  & Trial 4  & Trial 5  & Mean $\pm$ SD \\ \hlineB{3}
\multirow{2}{*}{FedAvg} & None                                                                                          & 0.8361          & 0.8308          & 0.8395          & 0.8409          & 0.8339          & $0.8363 \pm 0.0037$                             \\
                                 & SimCLR                                                                                        & 0.9062          & 0.8937          & 0.9022          & 0.9037          & 0.9097          & $0.9031 \pm 0.0054$                             \\ \hline
\multirow{2}{*}{IFCA}   & None                                                                                          & 0.9584          & 0.9586          & 0.9585          & 0.9605          & 0.9578          & $0.9587 \pm 0.0009$                             \\
                                 & FedAvg                                                                                        & 0.9646          & 0.9605          & 0.9624          & 0.9625          & 0.9631          & $0.9626 \pm 0.0013$                             \\ \hline
CP-CFL                  & SimCLR                                                                                        & \textbf{0.9778} & \textbf{0.9774} & \textbf{0.9786} & \textbf{0.9782} & \textbf{0.9779} & \bm{$0.9780 \pm 0.0004$}                    \\ \hlineB{3}
\end{tabular}}

\caption{AUROC (OvO macro) on multiple trials of the STL-10-to-STL-10 task ($T=100$). OvO, one-vs-one. SD, standard deviation.}
\label{tab:app_rocauc_ovomacro}
\end{table}

\begin{table}[htbp]
\centering
\resizebox{\linewidth}{!}{%

\begin{tabular}{l|lV{3}ccccc|c}
\hlineB{3}
                                 & \multicolumn{1}{cV{3}}{\multirow{2}{*}{\begin{tabular}[c]{@{}c@{}}Pre-\\ training\end{tabular}}} & \multicolumn{5}{c|}{AUROC (OvO weighted)}                                    &                                               \\
\textbf{}                        & \multicolumn{1}{cV{3}}{}                                                                                  & Trial 1  & Trial 2  & Trial 3  & Trial 4  & Trial 5  & Mean $\pm$ SD \\ \hlineB{3}
\multirow{2}{*}{FedAvg} & None                                                                                          & 0.8325          & 0.8285          & 0.8364          & 0.8384          & 0.8316          & $0.8335 \pm 0.0035$                             \\
                                 & SimCLR                                                                                        & 0.9046          & 0.8929          & 0.9000          & 0.9025          & 0.9084          & $0.9017 \pm 0.0052$                             \\ \hline
\multirow{2}{*}{IFCA}   & None                                                                                          & 0.9524          & 0.9529          & 0.9531          & 0.9552          & 0.9520          & $0.9531 \pm 0.0011$                             \\
                                 & FedAvg                                                                                        & 0.9598          & 0.9553          & 0.9572          & 0.9575          & 0.9580          & $0.9576 \pm 0.0014$                             \\ \hline
CP-CFL                  & SimCLR                                                                                        & \textbf{0.9747} & \textbf{0.9746} & \textbf{0.9758} & \textbf{0.9755} & \textbf{0.9750} & \bm{$0.9751 \pm 0.0005$}                    \\ \hlineB{3}
\end{tabular}}

\caption{AUROC (OvO weighted) on multiple trials of the STL-10-to-STL-10 task ($T=100$). OvO, one-vs-one. SD, standard deviation.}
\label{tab:app_rocauc_ovoweighted}
\end{table}

\newpage

\section{ResNet-18 encoder}
\label{sec:app_resnet_18}

Table~\ref{tab:app_resnet18_encoder_t} presents the evaluation results for the STL-10-to-STL-10 task obtained with the ResNet-18 encoder at $T=100$.

\begin{table}[htbp]
\centering

\begin{tabular}{l|lV{3}ccccc|c}
\hlineB{3}
\textbf{}                        & \multicolumn{1}{cV{3}}{\multirow{2}{*}{\begin{tabular}[c]{@{}c@{}}Pre-\\ training\end{tabular}}} & \multicolumn{5}{c|}{Accuracy (\%)}                                        &                                      \\
\textbf{}                        & \multicolumn{1}{cV{3}}{}                                                                                  & Trial 1   & Trial 2   & Trial 3   & Trial 4   & Trial 5   & Mean $\pm$ SD \\ \hlineB{3}
\multirow{2}{*}{FedAvg} & None                                                                                          & 24.67          & 34.00          & 25.10          & 27.53          & 35.70          & $29.40 \pm 4.59$                     \\
                                 & SimCLR                                                                                        & 34.83          & 29.40          & 33.83          & 30.73          & 29.67          & $31.69 \pm 2.22$                     \\ \hline
\multirow{2}{*}{IFCA}                    & None                                                                                          & 63.73          & 61.90          & 62.50          & 65.53          & 63.27          & $63.39 \pm 1.24$                     \\
                        & FedAvg                                                                                        & 63.63          & 63.17          & 64.80          & 62.90          & 61.70          & $63.24 \pm 1.01$                     \\ \hline
CP-CFL                  & SimCLR                                                                                        & \textbf{67.90} & \textbf{66.90} & \textbf{68.30} & \textbf{66.30} & \textbf{66.43} & \bm{$67.17 \pm 0.80$}            \\ \hlineB{3}
\end{tabular}
\caption{Test accuracy (\%) obtained with the ResNet-18 encoder on the STL-10-to-STL-10 task (${T=100}$). SD, standard deviation.}
\label{tab:app_resnet18_encoder_t}
\end{table}

\section{ResNet-50 encoder}
\label{sec:app_resnet_50}

Table~\ref{tab:app_resnet50_encoder} presents the evaluation results obtained with the ResNet-50~\citep{he2016deep} encoder on the STL-10-to-STL-10 task. For the contrastive pre-training step, we set the number of epochs to 500 and the batch size to 250. We use a projection head $\theta_g$ containing a ReLU-activated dense layer with 1024 neurons, followed by an another dense layer with 2048 neurons.
The output dimension of the ResNet-50 encoder is 2048, and we use $\theta_{c\text{-}2}$ as the classifier head.
Using the ResNet-50 encoder, CP-CFL can still outperform the baseline approaches as shown in Table~\ref{tab:app_resnet50_encoder}.

\begin{table}[htbp]
\centering

\begin{tabular}{l|lV{3}cc}
\hlineB{3}
\textbf{}                        & \multicolumn{1}{cV{3}}{\multirow{2}{*}{\begin{tabular}[c]{@{}c@{}}Pre-\\ training\end{tabular}}} & \multicolumn{2}{c}{Accuracy (\%)} \\
\textbf{}                        & \multicolumn{1}{cV{3}}{}                                                                                  & $T$=100              & Best                \\ \hlineB{3}
\multirow{2}{*}{FedAvg} & None                                                                                          & 45.10                & 45.80               \\
                                 & SimCLR                                                                                        & 39.07                & 46.13               \\ \hline
IFCA                    & None                                                                                          & 67.93                & 67.93               \\
\textbf{}                        & FedAvg                                                                                        & 69.57                & 71.10               \\ \hline
CP-CFL                  & SimCLR                                                                                        & \textbf{72.93}       & \textbf{72.93}      \\ \hlineB{3}
\end{tabular}
\caption{Test accuracy (\%) obtained with the ResNet-50 encoder on the STL-10-to-STL-10 task.}
\label{tab:app_resnet50_encoder}
\end{table}

\section{Pre-training time}
\label{sec:app_pretraining_costs}

We performed all of our experiments on a single RTX 3080 GPU with 10 GB of memory. Table~\ref{tab:app_pretraining_costs} compares the training time per epoch for SimCLR pre-training using different encoder architectures on the STL-10 unlabeled portion. We also calculate the computational cost of each encoder in flops using the \texttt{keras-flops}~\citep{tokusumi-2020} package. According to Table~\ref{tab:app_pretraining_costs}, pre-training with the CNN encoder takes approximately 2~h for 300 epochs, while pre-training with the ResNet-18 and ResNet-50 encoders takes around 8 and 18~h, respectively, for 500 epochs on our hardware.

\begin{table}[htb]
\centering

\begin{tabular}{lV{3}c|c}
\hlineB{3}
Encoder   & Time per epoch (s) & Flops (G) \\ \hlineB{3}
CNN       & $\approx 23$                                                                                 & 0.17               \\
ResNet-18 & $\approx 60$                                                                                 & 0.67               \\
ResNet-50 & $\approx 134$                                                                                & 1.42               \\ \hlineB{3}
\end{tabular}
\caption{Training time per epoch for SimCLR pre-training on the unlabeled data portion of the STL-10 dataset. We also report the computational cost of each encoder in flops.}
\label{tab:app_pretraining_costs}
\end{table}


\end{document}